\pdfoutput=1

\documentclass[11pt]{article}

\usepackage[final]{EMNLP2022}

\usepackage{times}
\usepackage{latexsym}
\usepackage{amsmath}
\usepackage{amssymb}
\usepackage{amsfonts}
\usepackage{booktabs}
\usepackage{subcaption}
\usepackage{graphicx}
\usepackage{bbm}
\usepackage{multirow}
\usepackage{comment}
\usepackage{microtype}

\usepackage[all]{hypcap}
\usepackage[nameinlink]{cleveref}
\usepackage{tikz}
\usetikzlibrary{patterns}
\usetikzlibrary{positioning}
\usetikzlibrary{decorations.markings}
\usetikzlibrary{decorations,arrows,shapes}
\usepackage{tikz-qtree}
\usepackage{color}
\usepackage[T1]{fontenc}

\usepackage[utf8]{inputenc}
\usepackage{framed}

\usepackage{microtype}

\usepackage{inconsolata}

\newcommand{\Pgen}{P_c}
\newcommand{\Pdata}{P_{\text{data}}}
\newcommand{\Pmodel}{P_{\text{model}}}
\newcommand{\Pagg}{P_{\text{agg}}}
\newcommand{\E}{\mathbb{E}}

\newcommand\mybox[2][]{\tikz[overlay]\node[fill=blue!20,inner sep=2pt, anchor=text, rectangle, rounded corners=1mm,#1] {#2};\phantom{#2}}
\newcommand\myboxtwo[2][]{\tikz[overlay]\node[fill=red!20,inner sep=2pt, anchor=text, rectangle, rounded corners=1mm,draw=red!80, line width=1mm,#1] {#2};\phantom{#2}}

\title{Model Criticism for Long-Form Text Generation}

\author{Yuntian Deng$^1$, Volodymyr Kuleshov$^2$, Alexander M. Rush$^2$\\\\
\vspace{3pt}
$^1$\:Harvard University \texttt{dengyuntian@seas.harvard.edu} \\
$^2$\:Cornell University \texttt{\{kuleshov,arush\}@cornell.edu} \\
}


\begin{document}
\maketitle
\begin{abstract}
Language models have demonstrated the ability to generate highly fluent text; however, it remains unclear whether their output retains coherent high-level structure (e.g., story progression). Here, we propose to apply a statistical tool, \emph{model criticism in latent space}, to evaluate the high-level structure of the generated text. Model criticism compares the distributions between real and generated data in a latent space obtained according to an assumptive generative process. Different generative processes identify specific failure modes of the underlying model. We perform experiments on three representative aspects of high-level discourse---coherence, coreference, and topicality---and find that transformer-based language models are able to capture topical structures but have a harder time maintaining structural coherence or modeling coreference.
\end{abstract}
\section{Introduction}
It is now broadly accepted that neural language models can consistently generate fluent text  
\citep{radford2019language,shoeybi2019megatron,brown2020language,smith2022using}. 
Yet, while large language models make few local word-level errors, human studies have shown that they still often make ``high-level'' errors such as incoherence, self-contradictions, and off-topic generations \citep{dou-etal-2022-gpt}. We hypothesize that researchers have focused on local fluency partly because it is easy to automatically evaluate through metrics such as perplexity and n-gram matching. Automatic assessment of high-level text generation quality has received less attention, partially because a single general-purpose metric does not exist. 

This work takes a step toward the automatic evaluation of 
the high-level structure of the generated text
by applying a tool from statistics, \emph{model criticism in latent space} \citep{dey1998simulation,seth2019model}. Under this approach, we first project data to a latent space based on an assumptive generative process, and then compare the implied latent distributions between real data and language model samples. 
This approach unifies past work for evaluating text generation under a single framework,
including existing dimensionality reduction techniques such as probabilistic PCA \citep{wold1987principal}, as well as previous applications of model criticism that were restricted to topic models \citep{mimno-blei-2011-bayesian}.

By making different assumptions in the underlying generative process, model criticism in latent space identifies specific failure modes of the generated language. 
We demonstrate this on three representative high-level properties of the generated discourse---coherence \citep{barzilay-lapata-2005-modeling}, coreference \citep{chomsky1993lectures}, and topicality \citep{blei2006correlated}---as well as on a synthetic dataset 
for which the true data generating process is known. 

Experiments using our proposed framework enable us to make four observations about modern language models.
First, we find that it is possible for a model to get strong word-level perplexity, yet fail to capture longer-term dynamics. 
Second, we find that the 
transformer language models perform poorly in terms of coherence, in line with previous observations \citep{dou-etal-2022-gpt,sun-etal-2021-long,krishna2022rankgen,sun-etal-2022-chapterbreak}, particularly when they do not have access to explicit lexical markers in the context. 
Third, we show that transformer language models do not model coreference structures well. Last, we show that transformer language models can capture topical correlations \citep{blei2006correlated}. 
All results, data, and code are publicly available at \url{https://github.com/da03/criticize_text_generation}.

\begin{figure*}[!htp]
\centering

\includegraphics[width=\textwidth,trim={0 0.2cm 0 0.4cm},clip]{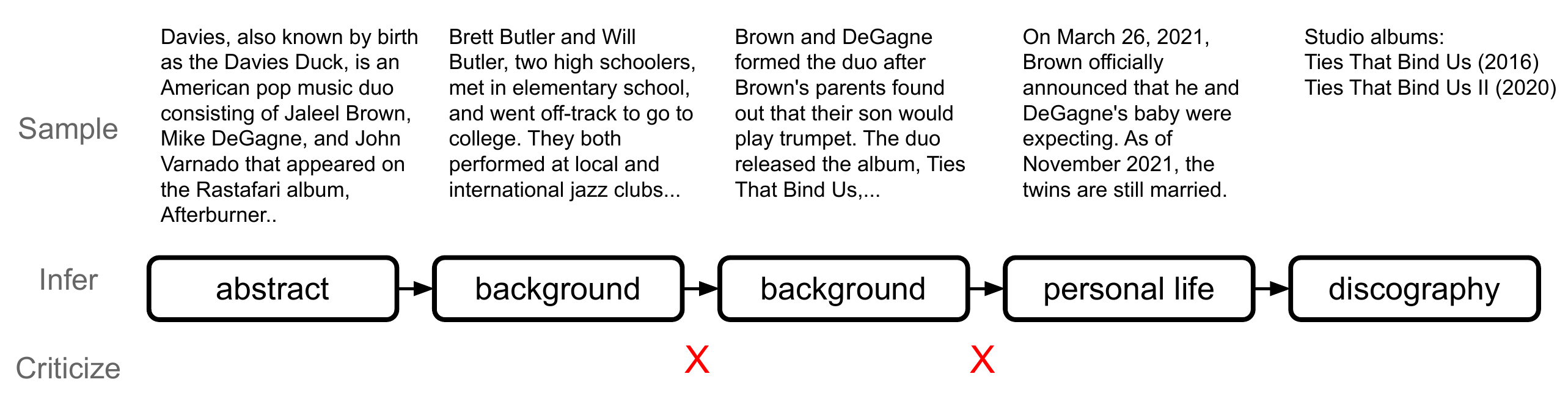}
\vspace{-0.6cm}
\caption{Illustration of applying model criticism in latent space to evaluate discourse coherence. Instead of word-level errors, we identify improper high-level section transitions (those that are rare in real data), as marked by red crosses. The article shown is generated by GPT-2 finetuned on \textsc{Wiki}. See \Cref{sec:dc} for more explanations.}
\label{fig:illustration_section}
\vspace{-0.25cm}
\end{figure*}

\section{Model Criticism in Latent Space\label{sec:mcls}}
%
Model criticism \citep{o2003hsss} quantifies the relationship between a data distribution $P_{\text{data}}(x)$ and a model $P_{\text{model}}(x)$ by comparing statistics over these two distributions. While model criticism can be applied to the observation space, in many applications we are interested in ``higher-level'' aspects of the data, such as the underlying topics of a document \citep{mimno-blei-2011-bayesian}, or the latent factors of an image \citep{seth2019model}. Model criticism in latent space \citep{dey1998simulation,seth2019model} lifts the criticism approach to a latent space in order to compute 
higher-level comparative statistics. 

\begin{figure}
    \centering
    \includegraphics[width=\linewidth,trim={0 1.1cm 0 1.1cm},clip]{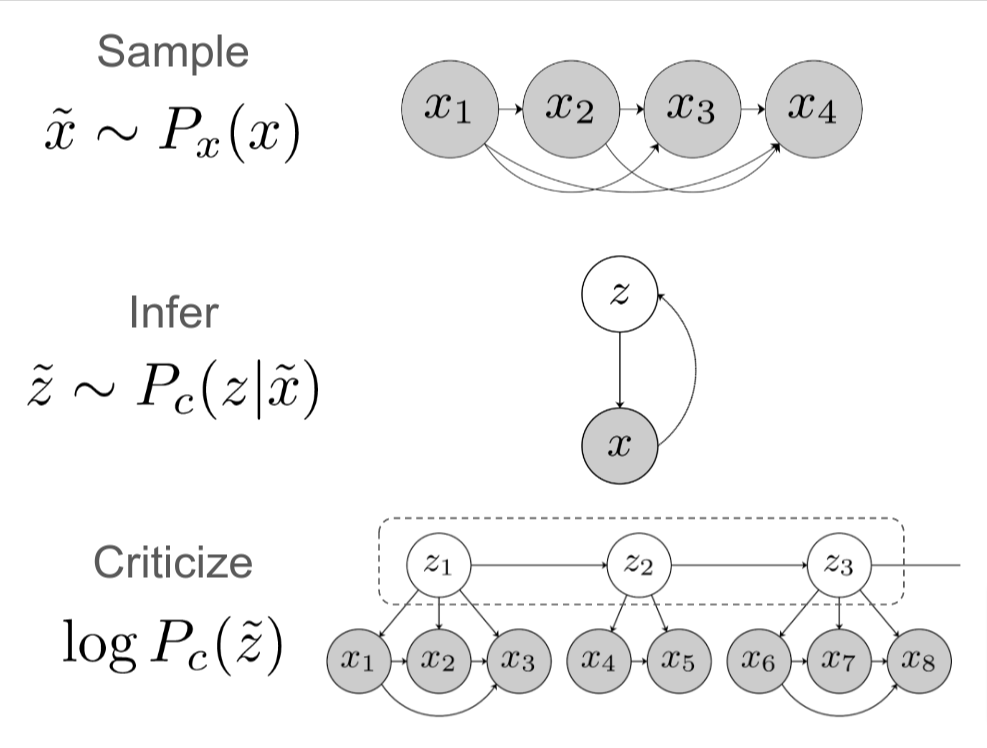}
    \vspace{-0.4cm}
    \caption{Model criticism in latent space. Given a sample $\tilde{x}$, we first map it to latent states $\tilde{z}$ using $\Pgen(z|\tilde{x})$. The likelihood of $\tilde{z}$ is evaluated using $\Pgen(\tilde{z})$ to measure how likely the samples are in the latent space.}
    \label{fig:crit}
\end{figure}

How do we critique latent properties of arbitrary, and perhaps unknown, distributions? For example, given a language model, how do we know how well it captures the section transitions at the discourse level (\Cref{fig:illustration_section})?
Lacking access to the generative process, we introduce a critic generative 
process $\Pgen$ with latent variables $z \in \mathcal{Z}$ and observations $x \in \mathcal{X}$:
\begin{align*}
    z \sim \Pgen (z) &&
    x \sim \Pgen(x|z).
\end{align*}
Based on this generative process, the posterior distribution $\Pgen(z|x)$ projects $x$ to the latent space. For a single data point $x$, we can evaluate the negative log-likelihood
of the projected latent variables $z\sim\Pgen(z|x)$ under the prior $\Pgen$,
\begin{align*}
T_c(x) &\triangleq -\mathbb{E}_{z\sim \Pgen(z|x)} \log \Pgen(z) \\
&= H(\Pgen(z|x), \Pgen(z)),
\end{align*}
where $H(p, q)\triangleq -\mathbb{E}_p \log q$ denotes the cross-entropy between two distributions $p$ and $q$.\footnote{We discuss the difference between being likely in the latent space versus the observed space 
in \Cref{sec:appendix_decomposition}.} This process is illustrated in~\Cref{fig:crit}.

Given an arbitrary distribution over $x$, $P_x$, we can take an expected negative log-likelihood,
$$
T_c(P_x) \triangleq -\mathbb{E}_{x\sim P_x(x)} \mathbb{E}_{z\sim \Pgen(z|x)} \log \Pgen(z).
$$

We term $T_c(P_x)$ the \emph{Latent NLL}.\footnote{When $z$ is the same as $x$ ($\Pgen(z|x)=\mathbbm{1}{[z=x]}$), Latent NLL is the same as the negative log-likelihood of the language model samples under the data distribution \citep{zhao2018adversarially}.} This value is the cross-entropy between the aggregated posterior distribution and the prior distribution of $z$:
$$
T_c(P_x) = H(\mathbb{E}_{x\sim P_x(x)} \Pgen(z|x), \Pgen(z)).
$$
In practice, we cannot compute $T(P_x)$ analytically due to the existence of the two expectations $\mathbb{E}_{x\sim P_x(x)}$ and $\mathbb{E}_{z\sim \Pgen(z|x)}$, but can approximate expectations using Monte-Carlo sampling. 

When $z$ is a sequence of $M$ discrete states, we define a metric \emph{Latent PPL} analogous to perplexity:
$$
\text{Latent PPL} (P_x) \triangleq \exp [T_c(P_x) / M].
$$

With a critic chosen, we can compare $P_{\text{data}}(x)$ and $P_{\text{model}}(x)$ in the latent space by estimating and comparing $T_c(P_{\text{data}})$ and $T_c(P_{\text{model}})$. Similar to a two-sample test \citep{hotelling1951generalized}, when $P_{\text{data}}$ and $P_{\text{model}}$ are the same, the statistics will also stay close. Furthermore, with a powerful critic, $T_c(\Pmodel)$ is meaningful by itself: a higher value means that model generations are less likely in the latent space, whereas a lower value implies that samples match the critic along the latent projection.\footnote{Under a bad critic, the value of $T_c(x)$ might not be meaningful even though we can still use it to compare distributions.} 
The approach can also be applied to individual points, $T_c(x)$, to identify outliers. 

\paragraph{How to select the critic $\mathbf{\Pgen}$} Choosing the critic $\Pgen$ is obvious only when we know the true latent variables and the generative process of data. In other cases, it depends on the data properties of interest. For example, if we want to criticize the topicality of text, we can use a topic model \citep{blei2003latent} to induce a latent space over topics. 
Note that the selected critic $\Pgen$ may underperform $\Pmodel$ as a model of $x$, while still providing useful latent structures. 
By criticizing strong models, using simpler latent models that are designed to capture a particular aspect of text as $\Pgen$, we provide a sanity check for the stronger model along a specific target axis. This property motivates the use of this approach with powerful, yet opaque models.

\section{\label{sec:latent}A Surprising Text Generation Failure}

As a preliminary experiment, we show a language model with strong word-level perplexity that fails to capture simple long-term dynamics as demonstrated by model criticism. 
We assume that $\Pdata$ is known and follows a basic pattern. 
It has a latent high-level sequence of $M=50$ discrete states $z_1, z_2, \ldots, z_M$ where each state can take one of 256 possible values. These states are generated from a transition distribution
$$
\Pdata(z_1, \ldots, z_M) = \prod_{m=1}^M\Pdata(z_m | z_{m-1}).\footnote{We assume a special beginning state $z_0$.}
$$
At the observation level, each latent state $z_m$ generates a sub-sequence of words $x_1^m, x_2^m, \ldots, x_{N_m}^m$  conditioned on $z_m$ from an emission distribution $P(x_1^m, \ldots, x_{N_m}^m | z_m)$. We also restrict the model so that each sub-sequence can only come from one latent state. 
The observed sequence is the concatenation of all sub-sequences. 
The joint distribution of the latent states and the tokens forms: 
\begin{align*}
    &\Pdata(x, z) = \Pdata(z) \Pdata(x|z)\\
    &=\prod_{m=1}^M\left [ \Pdata(z_m | z_{m-1}) \Pdata(x_1^m, \ldots, x_{N_m}^m | z_m) \right ].
\end{align*}


\noindent
With this generative process, we sample a dataset.\footnote{Sub-sequences vary between 4 to 11 words and the vocabulary size is set to 53. There are 51.2k samples for training, 6.4k for validation, and 6.4k for evaluation.}

We apply a transformer language model as $\Pmodel$ and train it on this dataset. Given the simplicity of the generative process and the small vocabulary size, we expect this model to do quite well. And in fact we do see that the model achieves a strong perplexity of 2.28, which nearly matches the true data $\Pdata$ perplexity of 1.99.

\begin{figure}[!t]
    Sample $x$: ... \mybox[fill=green!20]{p B W m <s>} \mybox[fill=blue!20]{T c g N f <s>} \\ \hspace*{1cm} \mybox[fill=red!20]{x i t K a b <s>} \mybox[fill=yellow!20]{b A x t N o m U <s>} ... \\
    Infer Latent $z$: ... \mybox[fill=green!20]{G}\ \mybox[fill=blue!20]{B}\ \mybox[fill=red!20]{R}\  \mybox[fill=yellow!20]{Y} ...\\
    Criticize (Latent NLL): ... \mybox[fill=green!20]{5.1} + \mybox[fill=blue!20]{9.9} + \mybox[fill=red!20]{10.3} + \mybox[fill=yellow!20]{4.2} ...
    \caption{Applying model criticism to synthetic data.}
    \label{fig:toy_example}
\end{figure}

\begin{table}[!t]
    \centering
    \begin{tabular}{@{}lcc@{}}
    \toprule
    & Trans-LM & HSMM-LM\\
    \midrule
    Word-level PPL & 2.28 & 2.05 \\
    \midrule
    Latent PPL (data) &  \multicolumn{2}{c}{44.30}  \\
    Latent PPL (model) & 64.80 & 47.24 \\
    \bottomrule
    \end{tabular}
    \caption{Evaluation results of transformer and HSMM on the synthetic dataset. Word-level PPL values are estimated on the test set, and Latent PPL values are estimated using the same number of samples (6.4k).
    }
    \label{tab:toy}
    \vspace{-0.2cm}
\end{table}

Model criticism gives a different method for quantifying model fit. Since the true data generating process is known, we can directly use $\Pgen = \Pdata$ as the critic to induce the latent space. To project an observation $x$ to the latent space, we need to perform posterior inference $\Pgen(z|x)$. By construction, this mapping is deterministic, since each sub-sequence comes from a unique latent state (see \Cref{sec:appendix_why_eval_latent} for details). We then apply model criticism $T_c$ by sampling a sequence of transformer outputs, 
mapping them to a sequence of latent states, counting to compute the aggregated posterior, and then comparing to the known prior. This process is shown in \Cref{fig:toy_example}.

\Cref{tab:toy} presents the results. Surprisingly, transformer gets a much worse Latent PPL compared to a hidden semi-Markov model (HSMM, the true model class) fit to data (66.80 v.s. 47.24), which has a near-optimal Latent PPL. This result implies that even though the transformer is nearly as good at predicting the next word in the sequence, it has not learned the higher-level transition structures. Seemingly, it can produce reasonable estimates of the next token which does not reflect the ability to capture longer-range dynamics of this system. 

\noindent \textbf{Motivation} Given this result, we ask whether similar issues are present in language models applied in more realistic scenarios. We therefore turn to experiments that consider model criticism for long-form generation, and ask whether language models capture properties of discourse coherence (\Cref{sec:dc}), coreference (\Cref{sec:coref}), and topicality (\Cref{sec:topic}).

\begin{table*}[!htp]
    \centering
    \begin{tabular}{@{}llcccccccc@{}}
    \toprule
    Metric & Model & \multicolumn{2}{c}{\textsc{PubMed}} & & \multicolumn{2}{c}{\textsc{ArXiv}} & & \multicolumn{2}{c}{\textsc{Wiki}} \\
    \cmidrule{3-4}\cmidrule{6-7}\cmidrule{9-10}
   &  & W/ Title & W/O Title & & W/ Title & W/O Title & & W/ Title & W/O Title\\
    \midrule
    \multirow{ 2}{*}{PPL}  & LM$_1$ & 11.38 & 11.50 & & 13.94 & 14.13 & & 15.38 & 15.84  \\
     & LM$_2$ & 10.96 & 11.09 & & 12.73 & 12.85 & & 16.35 & 16.86 \\
    \midrule
     \multirow{ 3}{*}{Latent PPL\ } & Data &  \multicolumn{ 2}{c}{2.58}   & & \multicolumn{ 2}{c}{3.87}  & & \multicolumn{ 2}{c}{4.80}\\
     & LM$_1$ & 2.68 & 3.76 &  & 6.72 & 9.52 & & 4.67  & 5.47\\
     & LM$_2$ &  4.17 & 7.92 & & 9.01 & 18.64 & & 6.48  & 10.22 \\
    \bottomrule
    \end{tabular}
    \caption{Results of coherence experiments. W/ Title is the setting where section titles are included in the training data for LMs, and W/O Title removes section titles from the training data.}
    \label{tab:main}
    \vspace{-0.2cm}
\end{table*}

\section{Critiquing Discourse Coherence\label{sec:dc}}



Text generation from large language models rarely leads to local fluency errors, but there is evidence of failures like those in the previous section~\citep{dou-etal-2022-gpt,sun-etal-2021-long,krishna2022rankgen,sun-etal-2022-chapterbreak}. In this section, we apply model criticism to assess discourse coherence~\citep{barzilay-lapata-2005-modeling} of large LMs. We study this through an experiment on generating long-form documents divided into explicit sections. While we do not know the true data generating process, knowing the distribution of section types allows us to assess the latent structure of LM generations.

\Cref{fig:illustration_section} illustrates the experiment. Here, an LM generates an article. Each word transition is fluent, but the system makes two section transition errors: first, it generates two sections of type ``background''; second, it generates a section of type ``personal life'' following the last ``background'' section, with both transitions being unlikely in the data.\footnote{``background'' is usually followed by ``reception''.} We aim to separate the evaluation of these high-level coherence errors from word-level errors.

To apply model criticism, we posit a simple critic generative process to capture the section changes. We adapt a hidden semi-Markov model (HSMM) which is commonly used to represent segmentations of this form. 
Specifically, the high-level latent variables $z_1,\ldots, z_M$\footnote{We prepend a special beginning state $z_0$ and append a special ending state $z_{M+1}$ that do not emit anything.} model transitions among section types and the bottom level generates text conditioned on the current section type:
\begin{align*}
    &\Pgen(x, z) = \Pgen(z) \Pgen(x|z)\\
    &=\prod_{m=1}^M\left [ \Pgen(z_m | z_{m-1}) \Pgen(x_1^m, \ldots, x_{N_m}^m | z_m) \right ].
\end{align*}

\begin{figure*}[!htp]
    \centering
    \includegraphics[trim={1cm 1cm 1cm 1cm},clip,width=0.99\linewidth]{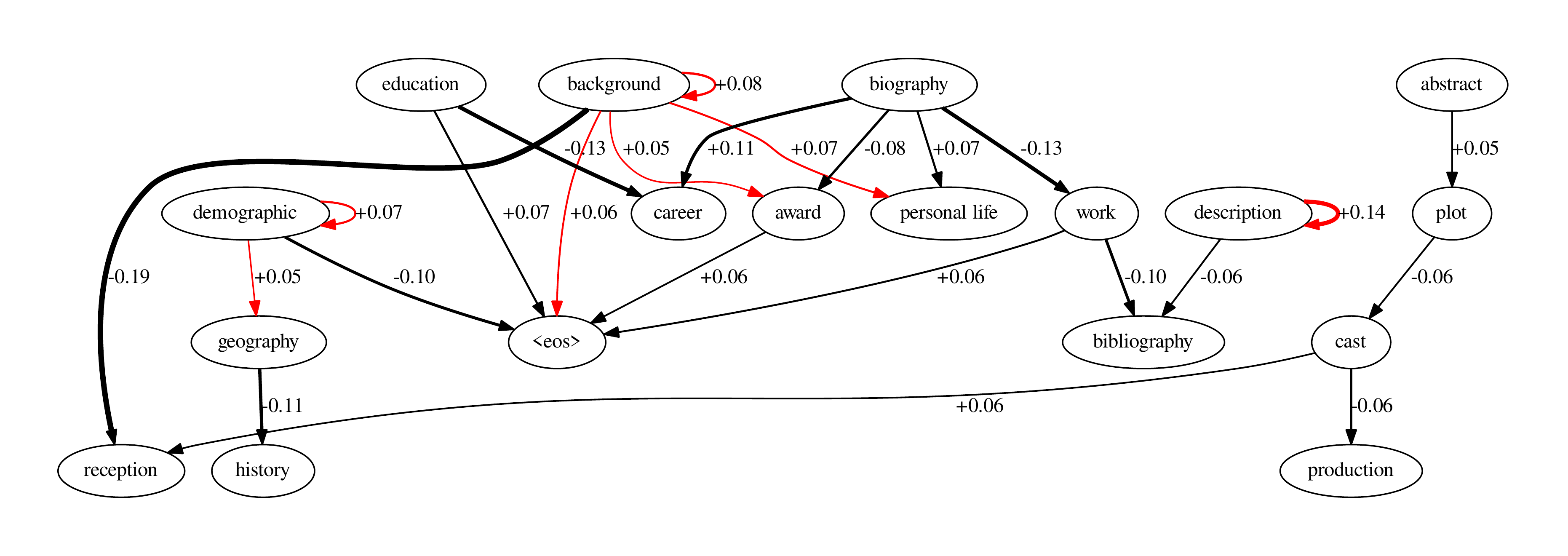}
\caption{Section transition errors on \textsc{Wiki}, where each edge is labeled with the difference between $P(z_m|z_{m-1})$ of LM$_1$ (W/O Title) and of data, and its width is proportional to the absolute difference. Red marks unlikely transitions ($\Pgen(z_m|z_{m-1})<0.05$). For clarity, we only show the top 20 section titles and remove singletons.\label{fig:dot}}
\vspace{-0.2cm}
\end{figure*}

We can then evaluate on datasets with known (ground truth) section titles and use these section titles as $z$. We use three English datasets \textsc{PubMed}, \textsc{ArXiv}, and \textsc{Wiki}~\citep{cohan-etal-2018-discourse}.\footnote{We adapt \textsc{PubMed} and \textsc{ArXiv} by filtering out section titles with low frequency. We download and process Wikipedia to get a dataset of the same format as \citet{cohan-etal-2018-discourse}.} 
We compare two language modeling settings, one trained with all section titles removed (``W/O Title'') and one with section titles before each section (``W/ Title''), since we hypothesize that the existence of explicit section type markers might help the model learn the dynamics, inspired by \citet{nye2021show} and \citet{wei2022chain}. Sections are separated by a special marker, and a special end-of-sequence symbol is used to mark the end of the generation. 
Since all three datasets are relatively small (especially considering that we use them to generate entire articles), we leverage pretrained language models GPT-2 small (LM$_1$) \citep{radford2019language}, and GPT-Neo small (LM$_2$) \citep{gpt-neo} which is trained on a more diverse dataset \citep{gao2020pile}. We finetune these LMs for $\Pmodel$.



    
    
    

    
    
    

To generate, we sample from the language model until we hit the end-of-sequence symbol. No tempering/truncation \citep{holtzman2019curious} is used during sampling, since we are more interested in the learned distribution rather than its mode here. 
For the ``W/ Title'' setting, we discard the generated section titles in a postprocessing step.

To infer the section types for a generated article, we need to approximate posterior inference to compute $T_c$. 
We make a simplifying assumption that the posterior section title of each section only depends on its corresponding text: $\Pgen(z |x) \approx \prod_{m=1}^M \Pgen(z_m | x_\cdot^m)$. We then finetune BERT with a classification head to estimate $\Pgen(z_m|x_\cdot^m)$. At inference time we use the MAP estimate of $z$ instead of maintaining the full distribution $P(z|x)$ (BERT is mostly over 90\% certain about its predictions). More details can be found in \Cref{sec:appendix_section}.

\begin{table}[!t]
    \centering
    \begin{tabular}{@{}llcc@{}}\toprule
        Metric &  Model & W/ Title & W/O Title \\
         \midrule
    \multirow{ 3}{*}{Latent PPL} & Data  & \multicolumn{ 2}{c}{4.78} \\
    & LM$_1$ & 5.18 & 6.24\\
    & LM$_2$ & 6.54 & 10.72 \\
    \bottomrule
    \end{tabular}
    \caption{Latent PPLs on \textsc{Wiki-Short} where all section transitions fit within the context window size.}
    \label{tab:main_short}
    \vspace{-0.35cm}
\end{table}

\paragraph{Results} \Cref{tab:main} gives results on coherence experiments. We first note that both models have strong word-level perplexity across datasets, with LM$_2$ doing better on two of the three datasets. We also note that removing titles has a negligible impact on the perplexity of the models. However, Latent PPL tells a different story. We find that LM$_1$ greatly outperforms LM$_2$ when criticizing with respect to the latent sections.\footnote{For one dataset, LM$_1$ has a lower Latent PPL than the data distribution. This result is possible as a consequence of our cross-entropy formulation of $T_c$, under which a mode-seeking distribution can get a lower value than $\Pdata$, which is the approximate entropy of the latent prior. }
It is also interesting that transformer LMs are sensitive to title words being explicitly included in the training data (i.e., the W/ Title setting). For example, LM$_1$~W/ Title gets a Latent PPL of 6.72 on \textsc{Arxiv}, whereas LM$_1$~W/O Title gets a Latent PPL of 9.52, despite having very close word-level PPLs (13.94 v.s. 14.13).
 These observations indicate that lacking explicit markers, the tested transformer LMs do not learn the long-term dynamics necessary for discourse coherence. Using explicit section topic markers might serve a similar functionality as using chain-of-thought prompting in language-model-based question answering tasks \citep{wei2022chain}.

One concern is that the difference between W/ Title and W/O Title is a side effect of language models having a limited context window size (1024 for LM$_1$ and 2048 for LM$_2$), since two adjacent sections might not fit within the context window size (but one section and the next section title are more likely to fit). To check if this is the case, we filter \textsc{Wiki} to only include articles with maximum section length 500 to form a new dataset \textsc{Wiki-Short}. In this dataset, any two adjacent sections can fit within the context window of both LM$_1$ and LM$_2$. \Cref{tab:main_short} shows that even in this case W/ Title still outperforms W/O Title, indicating that the difference between W/ Title and W/O Title is not due to the limited context window size.

\begin{table}[!t]
    \centering
    \small
    \begin{tabular}{@{}l@{}ccccc@{}}
    \toprule
        Metric & LM$_2$~W/O & LM$_2$~W/ & W/O & W/ & Data \\
        \midrule
        PPL$\downarrow$ & 11.09 & 10.96 & 11.50 & 11.38 & - \\
        MAUVE$\uparrow$ & 0.75 & 0.85 & 0.91 & 0.90 & 0.96 \\
        Latent PPL$\downarrow$ & 7.92 & 4.17 & 3.76 & 2.68 & 2.58 \\
        Human$\uparrow$ & 0.50     & 0.66     & 0.71 & 0.88 & 0.87 \\
         \bottomrule
    \end{tabular}
    \caption{Experiments on the correlation of Latent PPL (coherence) with human judgment and automatic metrics (\textsc{PubMed}). Latent PPLs agree well with human judgments of coherence. W/O: LM$_1$~W/O Title. W/: LM$_1$~W/ Title. See \Cref{sec:appendix_human} for evaluation details.}
    \label{tab:human_eval}
    \vspace{-0.35cm}
\end{table}

\Cref{fig:dot} visualizes the section transition errors made by LM$_1$ (W/O Title) for the most common section types on \textsc{Wiki}. We can find that the language model tends to generate the same section topic repeatedly, although there are other transition errors as well. More detailed error analysis can be found in \Cref{sec:appendix_section}.

\Cref{tab:human_eval} correlates automatic metrics with human judgments of coherence. Each human annotator first labels the section title of each section (after a training phase where they labeled and received feedback on real data), and then labels whether the organization of the section titles makes sense \citep{persing-etal-2010-modeling}. The baseline MAUVE \citep{pillutla2021mauve} is a metric that compares the distribution of GPT-3 hidden states between real data and model generations. From this table, we can observe that both MAUVE and Latent PPL align much better with humans than PPLs. 
Comparing MAUVE and Latent PPL, we can see that Latent PPL aligns better with humans: LM$_1$~W/O Title is considered to be better than LM$_1$~W/ Title under MAUVE, but both human evaluation and Latent PPL consider LM$_1$~W/ Title to be much better. 

\begin{table}[!t]
    \centering
    \begin{tabular}{@{}llcc@{}}\toprule
        Metric &  Model & W/ Title & W/O Title \\
        \midrule
    \multirow{4}{*}{PPL}& GPT-2 S & 15.38 & 15.84\\
    & GPT-2 M & 12.98 & 13.32 \\
    & GPT-2 L & 12.19 & 12.52 \\
    & GPT-2 XL & 11.60 & 11.99 \\
         \midrule
    \multirow{5}{*}{Latent PPL} & Data  & \multicolumn{ 2}{c}{4.80} \\
    & GPT-2 S & 4.67 & 5.47 \\
    & GPT-2 M & 4.79 & 5.58 \\
    & GPT-2 L & 4.90 & 5.75 \\
    & GPT-2 XL & 4.75 & 5.56 \\
    \bottomrule
    \end{tabular}
    \caption{The results of scaling model size on \textsc{Wiki}. Increasing model size improves PPL but not Latent PPL.}
    \label{tab:section_scaling}
    \vspace{-0.2cm}
\end{table}

A natural question is whether increasing model size improves coherence. To this end, in addition to GPT-2 small (GPT-2 S, aka LM$_1$, 117M parameters), we apply model criticism to GPT-2 medium (GPT-2 M, 345M parameters), GPT-2 large (GPT-2 L, 742M parameters), and full GPT-2 (GPT-2 XL, 1.5B parameters) on \textsc{Wiki}. The results are summarized in~\Cref{tab:section_scaling}. We can see that increasing model size improves PPL but not Latent PPL.

\section{Critiquing Coreference Chains\label{sec:coref}}

\begin{figure}[!t]
    \begin{framed}
    \textbf{Original Text}
    
    ... \mybox[fill=green!20]{[Lisa]$_0$} runs off to find \mybox[fill=blue!20]{[him]$_1$} and \mybox[fill=yellow!20]{[they]$_2$} kiss passionately. Afterwards, \mybox[fill=blue!20]{[Josh]$_1$} tells \mybox[fill=green!20]{[her]$_0$} the reason why \mybox[fill=blue!20]{[he]$_1$}'s going to \mybox[fill=yellow!20]{[their]$_2$} first gig, and that \myboxtwo[fill=green!20]{[Lisa]$_0$} is going to do it, too...\\
    
    \textbf{Coreference Chains} $\mathbf{z}$
    
    . \mybox[fill=green!20]{[Female]$_0$} \mybox[fill=blue!20]{[him]$_1$} \mybox[fill=yellow!20]{[they]$_2$} . \mybox[fill=blue!20]{[Male]$_1$} \mybox[fill=green!20]{[her]$_0$} \mybox[fill=blue!20]{[he]$_1$} \mybox[fill=yellow!20]{[their]$_2$} \myboxtwo[fill=green!20]{[Female]$_0$}\\
    
    \textbf{5-gram critic} $\mathbf{P_c}$
    \begin{align*}
        &P_c(\ \myboxtwo[fill=green!20]{\text{[Female]$_0$}}\ \ | \  \text{previous entity mentions}) \\
        &\approx P_c(\ \myboxtwo[fill=green!20]{\text{[Female]$_0$}}\ \  |\ \  \mybox[fill=blue!20]{\text{[Male]$_1$}} \mybox[fill=green!20]{\text{[her]$_0$}} \mybox[fill=blue!20]{\text{[he]$_1$}} \mybox[fill=yellow!20]{\text{[their]$_2$}})
    \end{align*}
    \end{framed}
    \caption{Critiquing coreference chains on a sample from LM$_1$. We first extract entity mentions from the text and only keep the genders of proper nouns to form $\mathbf{z}$, then a 5-gram $P_c$ is used to score $\mathbf{z}$. []$_i$ denotes a mention with entity id $i$. \textbf{.} marks sentence boundaries. 
    }
    \label{fig:coref}
    \vspace{-0.3cm}
\end{figure}

Coreference tracks how multiple \textit{mention} expressions are used to refer to the same underlying \textit{entity} \citep{karttunen-1969-discourse-referents,gordon1998representation}. While coreference represents a ubiquitous and important discourse-level phenomenon \citep{jurafskyspeech, kunz-hardmeier-2019-entity}, there is evidence that large neural language models make elementary coreference mistakes \citep{pagnoni-etal-2021-understanding}, such as referring to non-existent discourse entities \citep{schuster-linzen-2022-sentence}.

In this experiment, we compare the coreference chain \citep{jurafskyspeech} distributions between real data and LM generations. A coreference chain consists of a sequence of coreferent mentions. To simplify the representation, we use gender features to replace non-pronominal tokens, as illustrated in \Cref{fig:coref}.\footnote{We discuss the ethical considerations of using gender as features in \Cref{sec:ethical}.} Presumably these latent chains should be similar in generated text and in real data.

For the critic $P_c$, we use a 5-gram language model with Kneser–Ney smoothing \citep{ney1994structuring} over chains. To infer $\mathbf{z}$, we use an off-the-shelf coreference resolution tool.\footnote{\url{https://github.com/huggingface/neuralcoref}} To avoid data sparsity issues, we relabel entity clusters within each n-gram. We apply model criticism to compare real data and LMs trained on \textsc{Wiki} (W/ Title), after filtering data to only consider articles about films since they contain richer reference structures. 

\begin{table}[t]
    \centering
    \begin{tabular}{@{}lllllcc@{}}
    \toprule
       $z_{1}$ & $z_2$ & $z_3$ & $z_4$ & $z_5$ & \hspace{-0.3cm}$\hat{P}_{\text{data}}$  & $\hat{P}_{\text{LM}}$ \\
        \midrule
         &  &  & . & M$_0$ & 0.10 & 0.12 \\\relax
        . & M$_0$ & . & M$_0$ & M$_0$ & 0.01 & 0.03 \\\relax
        . & . & M$_0$ & his$_0$ & he$_0$ & 0.04 & 0.05 \\\relax
        . & . & . & N$_0$ & His$_1$ & 0.00 & 0.00$^\dagger$\hspace{-0.13cm} \\\relax
        M$_0$ & M$_1$ & . & M$_0$ & M$_0$ & 0.00 & 0.01 \\\relax
        N$_0$ & N$_1$ & M$_2$ & M$_3$ & We$_4$ & 0.00 & 0.00$^\dagger$\hspace{-0.13cm}  \\\relax
        . & F$_0$ & her$_0$ & . & F$_0$ & 0.03 & 0.04 \\\relax
        M$_0$ & his$_0$ & . & M$_1$ & M$_1$ & 0.00$^\dagger$\hspace{-0.13cm} & 0.01 \\\relax
        . & . & N$_0$ & N$_1$ & h.self$_2$ & 0.00 & 0.00$^\dagger$\hspace{-0.13cm} \\\relax
        . & . & M$_0$ & his$_0$ & M$_0$ & 0.01 & 0.01 \\
         \bottomrule
    \end{tabular}
    \caption{Coreference chain n-grams ranked by contribution to the difference in Latent NLL (LM$_1$). $\hat{P}$ denotes the empirical frequency of an n-gram in percentages.  
    M (Male), F (Female), and N (None of the above). Blank: padding. 
    $^\dagger$: small positive numbers truncated to 0.} \label{tab:result_coref_truecontribution}
\end{table}

\begin{table}[!t]
    \centering
    \begin{tabular}{@{}ccc@{}}
    \toprule
       Data & LM$_1$ & LM$_2$ \\
        \midrule
       6.26 & 7.22 &  6.93 \\
        \bottomrule
    \end{tabular}
    \caption{Latent PPLs of coreference chains.}
    \label{tab:latent_coref}
\end{table}

\paragraph{Results} \Cref{tab:latent_coref} shows the Latent PPLs on real data and LM generations. We can see that in general there is a mismatch of coreference distributions. Interestingly, while LM$_1$ models outperformed LM$_2$ models on discourse coherence, for this task LM$_2$ models are better.  

\Cref{tab:result_coref_truecontribution} shows the 10 coreference chain n-grams that most contributed to this difference. Some are intuitively implausible: in the fourth row, [His]$_1$ does not have a local antecedent; in the second to the last row, [himself]$_2$ also does not have a local antecedent. Others are rare but possible: in the last row, a proper noun [Male]$_0$ is used after a pronoun [his]$_0$ is used in the same sentence to refer to the same entity.\footnote{Different from prescriptive linguistic theories on coreference \citep{chomsky1993lectures,buring2005binding}, the differences identified by model criticism only reflect differences in empirical distributions and do not necessarily mean coreference errors.}   

The learned critic $P_c$ can also be used to identify unlikely coreference chains, as shown in \Cref{tab:result_critic} in \Cref{sec:appendix_coref}. \Cref{sec:appendix_coref} also has more qualitative examples and analyses. 

Lastly, we evaluate whether scaling model size improves coreference modeling. The results are summarized in~\Cref{tab:coref_scaling}. We can see that increasing model size does not improve Latent PPL, similar to our observations on critiquing discourse coherence.

\begin{table}[!t]
    \centering
    \begin{tabular}{lc}\toprule
      Model & Latent PPL \\
        \midrule
   Data  & {6.26} \\
     GPT-2 S & 7.22 \\
     GPT-2 M & 7.64  \\
     GPT-2 L & 7.27 \\
     GPT-2 XL & 7.62 \\
    \bottomrule
    \end{tabular}
    \caption{Critiquing coreference chains on larger models. Increasing model size does not improve Latent PPL.}
    \label{tab:coref_scaling}
    \vspace{-0.2cm}
\end{table}

\begin{figure*}[t]
  \centering
  \includegraphics[width=\textwidth,trim={0cm 0cm 0cm 0cm},clip]{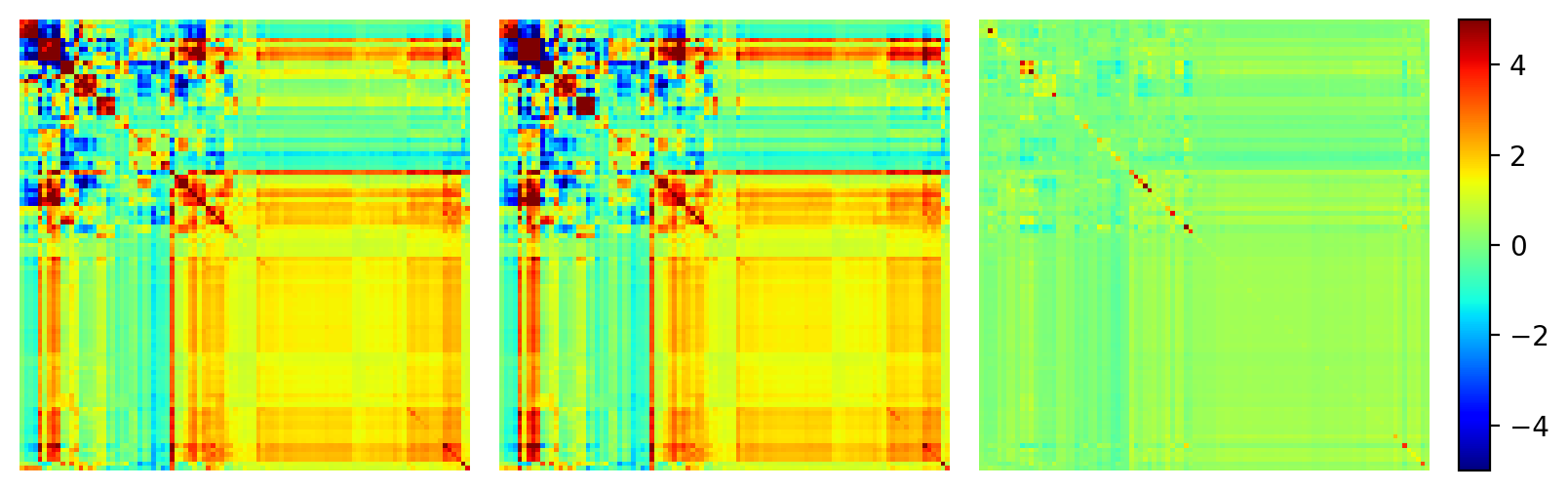}
  \caption{\label{fig:cov} Topic covariance matrix for the induced $z$ (on \textsc{Wiki}). Left: Test set ($\Pdata$). Middle:  LM$_1$ generations ($\Pmodel$). Right: generations of LM$_1$ trained on \textsc{PubMed} as a visual baseline. The Latent NLLs are: 124.70, 123.30, and 140.10. Topic ids are rearranged using hierarchical clustering to facilitate visual comparison.}
  \vspace{-0.2cm}
\end{figure*}

\section{Critiquing Topic Correlations\label{sec:topic}}

Topical structure is another important aspect of long-form document generation~\cite{serrano2009modeling}.
Certain topics are more likely to appear together, for example, a document containing a topic related to ``poets'' is more likely to also contain one related to ``publisher'' relative to one related to ``football''. A text generation model should capture these topical relations.
 For this experiment, we again sample documents from the trained language model $\Pmodel$. Specifically, we utilize the transformer-based LMs trained on the datasets in \Cref{sec:dc} (W/O Title).





To explore the topical structure in the generated documents, we need a critic $\Pgen$. While LDA \citep{blei2003latent} is the most commonly used generative process for topic modeling, the Dirichlet prior does not explicitly model topic correlations in documents. We therefore use the correlated topic model (CTM) specifically designed to model topical structures~\citep{blei2006correlated}. Model criticism will then compare the latent space of the real data with the generated texts.

For each document, a CTM with $M$ topics first generates a topic coefficient latent variable $z\in\mathbb{R}^{M}$ from a multivariate Gaussian distribution $\Pgen(z) \triangleq \mathcal{N}(z; \mu, \Sigma).$

Each coefficient of $z$ can be interpreted as the ``strength'' of a topic in a document, so the covariance matrix $\Sigma$ captures the correlations among different topics. These weights $z$ are then normalized using a softmax function, the result of which is used to parameterize the distribution over topic $t_n$ for the $n$-th word. Each topic $t_n$ induces a categorical distribution over word types $P(x_n|t_n)=\phi_{t_n, x_n}$, where $\phi_{ij}$ parameterizes the probability of emitting word type $j$ conditioned on topic $i$. The joint probability of a document with $N$ words is:
$$
\Pgen(x, t | z; \phi) = \prod_{n=1}^N{[\text{softmax}(z)]}_{t_n}\phi_{t_n, x_n}.
$$

Since we are only interested in criticizing the document-level $z$, we marginalize out the topic assignments of individual words:
$$
\Pgen(x|z) = \prod_{n=1}^N \sum_{i=1}^{M} {[\text{softmax}(z)]}_i \phi_{i,x_n}.
$$

To fit this generative process on data, we use variational inference and maximize the ELBO following \citet{blei2006correlated}. 
We set $M$ to 100. Since analytical posterior inference is intractable, we use variational inference to estimate $\Pgen(z|x)$.




\begin{table}[t]
    \centering
    \begin{tabular}{@{}ll@{ }c@{ }c@{ }c@{}}
    \toprule
        Metric & Model & \textsc{PubMed} & \textsc{ArXiv} & \textsc{Wiki}\\
        \midrule
         \multirow{ 3}{*}{Latent NLL} & Data & 174.43 & 163.95 & 124.70\\ 
        & LM$_1$ & 172.70 & 161.40 & 123.30\\ 
        & LM$_2$ & 172.81 & 163.17 & 124.35\\ 
        \bottomrule
    \end{tabular}
    \caption{Latent NLL of topic correlation modeling. Transformer LMs perform similarly to the real data.}
    \label{tab:ctm}
    \vspace{-0.25cm}
\end{table}


\paragraph{Results} \Cref{tab:ctm} shows the main results. The Latent NLLs of LM generations and real data are close on all three datasets (there are outlier pathological generations that we can identify using $T(x)$, as shown in \Cref{sec:appendix_topic}). In \Cref{fig:cov}, we visualize and compare the covariance matrices of the aggregated posterior distributions of LM generations and real data, and find that transformers are able to model the correlations among topics well. These results indicate that topic correlation is well represented in text generation systems, and is likely an easier task to model than ordered coherence.

\section{Related Work}

\paragraph{Text Generation Evaluation} 
Traditional evaluation metrics include perplexity, 
and n-gram overlap metrics for translation-type problems such as BLEU \citep{papineni-etal-2002-bleu}, ROUGE \citep{lin-2004-rouge}, METEOR \citep{lavie-agarwal-2007-meteor}, and NIST \citep{martin2000nist}.  In recent years, with the emergence of neural models that learn contextual representations \citep{devlin-etal-2019-bert,liu2019roberta}, researchers propose to project text to contextual representations and compute distance in this space \citep{zhang2019bertscore,zhao-etal-2019-moverscore,pillutla2021mauve}. The closest work to ours is \citet{eikema-aziz-2020-map}, which evaluates different decoding strategies in machine translation by comparing the statistics of the produced text. While these past works mainly concern word-level string/meaning representation matching, the goal of our work is to check the high-level aspects of the generated text such as coherence. Besides, word-level matching is not suitable for evaluating open-ended generation tasks due to the existence of too many plausible references 
\citep{celikyilmaz2020evaluation}, while our work projects text to a more manageable
lower-dimensional latent space to make the evaluation of open-ended generation feasible. 

\paragraph{Evaluation of Long-Form Text} 
There is a long line of research evaluating the discourse coherence of text \citep{grosz-etal-1995-centering,poesio-etal-2004-centering,barzilay-lapata-2005-modeling,lai-tetreault-2018-discourse,logeswaran2018sentence,persing-etal-2010-modeling}. 
Most learn a predictor that maps features such as the distribution of entities \citep{barzilay-lapata-2005-modeling} or the transitions of topics \citep{persing-etal-2010-modeling} to manually-labeled coherence scores. 
Our work differs in two important ways: first, we unify the evaluation of different high-level aspects of text using the formalism of model criticism; second, we do not assume any annotated coherence scores---we only specify a generative process in order to project text to a latent space for the comparison between machine-generated text and real text.  Recently, there have been works targeting the evaluation of discourse-level coherence, such as BARTScore \citep{yuan2021bartscore} and DiscoScore \citep{zhao2022discoscore}. These methods presume either a conditional generation setting or require textual references. We also note that model criticism does not use a generic neural representation, but focuses on specific user-specified high-level aspects of text. In this respect, our work is similar in spirit to some recently proposed suite-based metrics, such as Language Model Evaluation Harness~\citep{eval-harness} and BIG-bench~\citep{srivastava2022beyond} that utilize many different skill-based metrics. 

\paragraph{High-Level Issues of Text Generation Models} Concurrent with our work, several other groups also notice that existing LMs fail to capture some high-level aspects of text. For example, similar to our findings of LMs being not strong at discourse coherence, \citet{sun-etal-2022-chapterbreak} observe that large LMs including GPT-3 fail to assign a higher probability to ground truth chapter continuations compared to distractor chapters given a prefix, and yet a simple classifier trained on this identification objective can achieve a much higher accuracy.  Similar to our findings of LMs being not strong at modeling coreference,  \citet{pmlr-v162-papalampidi22a} find that LMs fail to maintain long-range entity consistency and coherency in the generated narrative stories.

\paragraph{Model Criticism}
Model criticism, also known as model checking, is a general framework for checking if a generative model fits the data well 
\citep{box1980sampling,gelman1995bayesian,stern2005bayesian,o2003hsss}. Model criticism is different from aforementioned metrics such as PPL and is similar to two-sample tests \citep{hotelling1951generalized} in that it computes and compares some statistics on the real data and on the samples to determine if they are close enough. While the statistics may be directly computed in the observation space, in many applications we are interested in criticizing some latent aspects of data such as topics \citep{mimno-blei-2011-bayesian} or latent factors \citep{seth2019model}. To this end, \citet{dey1998simulation} introduce model criticism in latent space, which measures the discrepancy between real data and model generations in the latent space induced by a generative model \citep{chaloner1988bayesian,o2003hsss,seth2019model,dey1995simulation,weiss1995residuals,dey1998simulation}.  
Recently, \citet{barkhof2022statistical} propose to use model criticism to evaluate VAEs. 
Model criticism in latent space forms the basis of our work, with two major differences: first, we apply model criticism to models with a point estimate of parameters such as commonly-used neural language models instead of models with uncertainties in their parameters. 
Second, we allow for using a different generative model to induce the latent space from the model that we criticize. By separating out the model to be criticized and the generative process used for projecting data to the latent space, our approach allows for criticizing different views of the data depending on user needs and for criticizing generative models without any latent variables such as neural language models. 
For qualitative analysis and outlier identification, our work applies visual posterior predictive checks \citep{gabry2019visualization,gelman1997bayesian}, a graphical version of model criticism.

\section{Limitations}
One limitation of the proposed approach is its reliance on choosing a critic generative process $\Pgen$, which presumes some knowledge of a true data generating process. For an improperly specified critic, it does not expose the latent space that we intend to criticize. However, since we compare statistics between real data and model generations (similar to two-sample tests), for a good model the statistics should be close even with improper critics.

Another limitation is that not observing any differences does not imply that the model generations conform to the unknown data distribution---it simply means that they are close with regard to the latent aspects that we criticize~\citep{o2003hsss}. 

Recently, researchers found that certain capabilities such as reasoning under augmented prompts only emerge in large LMs beyond tens of billions of parameters~\citep{wei2022emergent}. Since the largest LM tested in this paper only has 1.5 billion parameters, future work is required to investigate whether the high-level issues observed in this paper can be solved by further scaling model size.

\section{Conclusions}
 We consider the problem of evaluating long-form text generation for specific discourse properties. We propose a statistical tool, model criticism in latent space, which projects text to a latent space based on an assumptive generative process, and compares the implied latent distribution. Different critic generative processes focus on different properties of data. We apply this tool to analyze three representative document properties: coherence, coreference, and topicality, using transformer-based language models. Experiments find that while transformer LMs can capture topical structures well, they are not currently strong at modeling discourse coherence without explicit markers or at modeling coreference. 
 



\section{\label{sec:ethical}Ethical Considerations}
In our experiment of critiquing coreference chains, we used a gender binary \citep{hyde2019future} to categorize proper nouns, but there are many individuals who do not adhere to the gender binary that this simple categorization fails to consider \citep{bamman2014gender}. The reason for the gender binary is primarily because personal pronouns are typically gendered in English which makes the qualitative and statistical analysis more clear. For example, one coreference error detected by the approach is to use pronouns of different genders to refer to the same person, as shown in \Cref{sec:appendix_coref}. In \Cref{sec:appendix_coref}, we describe the exact procedure through which the genders of proper nouns are determined to make explicit what our ``gender'' definition is \citep{larson2017gender}. Going forward, exploring other features of proper nouns such as their syntactic features \citep{shieber-tao-2003-comma} to replace gender assignments here might further mitigate this concern.

\section*{Acknowledgements}
YD is supported by an Nvidia Fellowship. AR is
supported by NSF CAREER 2037519, NSF 1704834, and a Sloan Fellowship. We would also like to thank
Harvard University FAS Research Computing for providing computational resources.

\bibliography{anthology,custom}
\bibliographystyle{acl_natbib}
\clearpage
\section*{Appendix}
\appendix
\section{Interpretation of Latent NLL}
\label{sec:appendix_decomposition}
In \Cref{sec:latent} we termed $T(x)$ the Latent NLL, and a lower $T(x)$ indicates being more likely in the latent space. What does it mean to be ``more likely in the latent space''? How is it reflected in the marginal likelihood $P(x)$? In this section we answer this question by decomposing the log marginal likelihood $\log P(x)$ into three components (for brevity we use $P$ instead of $\Pgen$ in this section):\footnote{The decomposition is in fact the evidence lower bound (ELBO) where the expectation is taken w.r.t. the true posterior distribution, so the inequality becomes tight.}
\begin{align*}
    &\log P(x)\\
    &= \frac{\log P(x, z)}{\log P(z|x)}\ (\forall z) \text{ (Bayes' theorem)}\\
    &=  \mathop{\E}_{z \sim P(z|x)} \frac{\log P(x, z)}{\log P(z|x)}\\
    &= \mathop{\E}_{z \sim P(z|x)} \frac{\log P(x|z) + \log P(z)}{\log P(z|x)}\\
    &= \underbrace{\mathop{\E}_{P(z|x)} \log P(z)}_{\text{(1)}} + \underbrace{\mathop{\E}_{P(z|x)} \log P(x|z)}_{\text{(2)}}  + { \underbrace{\vphantom{ \mathop{\E}_{P(z|x)} \log P(z) }H(P(z|x))}_{\text{(3)}}}
\end{align*}

The first term (1) can be interpreted as how likely the posterior latent variable of $x$ is under the prior distribution $P(z)$ in expectation, and it is the negative Latent NLL ($-T(x)$) according to the definition of $T$. The second term (2) can be understood as how likely it is to realize the observation $x$ given the posterior latent codes $z$. The third term measures the diversity of the posterior distribution, which is not reflected in our evaluation metric. In fact, if we combine (1) and (3) we would get $-\text{KL}(P(z|x) || P(z))$: 
$$
\log P(x) = \mathop{\E}_{P(z|x)} \log P(x|z) -\text{KL}(P(z|x)||P(z)) 
$$

Therefore, the proposed evaluation metric $T(x)$ (hence $T(P_x)$) can be complemented using a diversity measure for completeness, which we leave for future work. 



\section{The Optimal Critic Prior $\Pgen(z)$}
\label{sec:appendix_optimal}
For the quantity $T(x)$ to be meaningful, $\Pgen(z)$ should not be an uninformative prior. For example, if $\Pgen(z)$ is uniform, then $T(x)$ hence $T(P_x)$ would be a constant. We will show below that the optimal $\Pgen(z)$ that maximizes the data likelihood is exactly the aggregated posterior distribution under the data distribution ($\Pagg(z)\triangleq\mathbb{E}_{x\sim P_{\text{data}}(x)} \Pgen(z|x)$). 



To find the optimal $\Pgen(z)$ that maximizes the data likelihood, we use the equation from \Cref{sec:appendix_decomposition} that $\log \Pgen(x)=\mathop{\E}_{\Pgen(z|x)} \log \Pgen(z) + \mathop{\E}_{\Pgen(z|x)} \log \Pgen(x|z) + H(\Pgen(z|x))$, and take the expectation on both sides w.r.t. $\Pdata(x)$:
\begin{align*}
    &\mathop{\E}_{x\sim \Pdata} \log \Pgen(x) \\
    &= \mathop{\E}_{x\sim \Pdata}\mathop{\E}_{\Pgen(z|x)} \log \Pgen(z) \\
    &\ \ \ \  + \mathop{\E}_{x\sim \Pdata}\mathop{\E}_{\Pgen(z|x)} \log \Pgen(x|z)\\
    &\ \ \ \  +  \mathop{\E}_{x\sim \Pdata} H(\Pgen(z|x))\\
    &= \mathop{\E}_{z \sim \Pagg} \log \Pgen(z) + \mathop{\E}_{x\sim \Pdata}\mathop{\E}_{\Pgen(z|x)} \log \Pgen(x|z)\\
    &\ \ \ \ +  \mathop{\E}_{x\sim \Pdata} H(\Pgen(z|x))\\
    & = -\text{KL}(\Pagg(z)||\Pgen(z)) +H(\Pagg(z)) \\
    &\ \ \ \ + \mathop{\E}_{x\sim \Pdata}\mathop{\E}_{\Pgen(z|x)} \log \Pgen(x|z)\\
    &\ \ \ \ +  \mathop{\E}_{x\sim \Pdata} H(\Pgen(z|x)).
\end{align*}

In the right-hand side of the above equation, the only term containing $\Pgen(z)$ is the first term $-\text{KL}(\Pagg(z)||\Pgen(z))$. Therefore, the optimal $\Pgen(z)$ that maximizes the likelihood of data is $\Pagg(z)$, although the optimization algorithm is not guaranteed to find this optimum.

At its optimality, $\Pgen(z)$ is the same as the aggregated posterior distribution $\Pagg(z)$, in which case $T(P_x)$ can also be interpreted as the cross-entropy between the aggregated posterior under model generations ($\E_{x\sim P_x(x)}\Pgen(z|x)$) and the aggregated posterior under the real data distribution ($\E_{x\sim \Pdata(x)}\Pgen(z|x)$):
$$
T(P_x) = H(\E_{x\sim P_x(x)}\Pgen(z|x), \Pagg)
$$

\section{Detection of Code-Mixing}
In this section, we show that model criticism can generalize some previous high-level evaluation metrics. In particular, we replicate the machine translation experiment in \citet{zhou2019understanding} under our framework. In this experiment, an English sentence might be translated into Spanish, German, or French, but never a mix of different languages. Therefore, one failure mode of a model is to generate text that contains code-mixing. 

\begin{figure*}[t]
  \centering
  \begin{subfigure}[b]{0.32\textwidth}
  \centering
  \includegraphics[height=3.5cm,trim={4cm 4cm 4cm 4cm},clip]{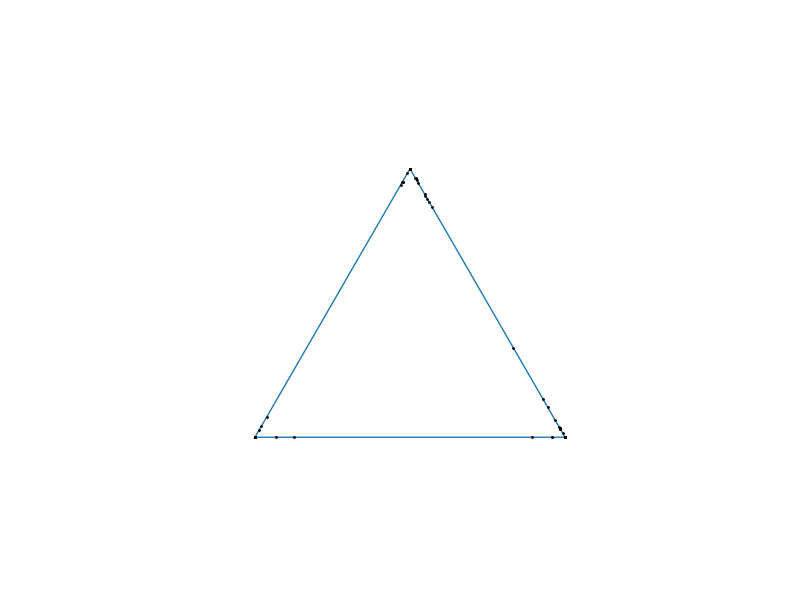} 
  \end{subfigure}
  \begin{subfigure}[b]{0.32\textwidth}
  \centering
  \includegraphics[height=3.5cm,trim={4cm 4cm 4cm 4cm},clip]{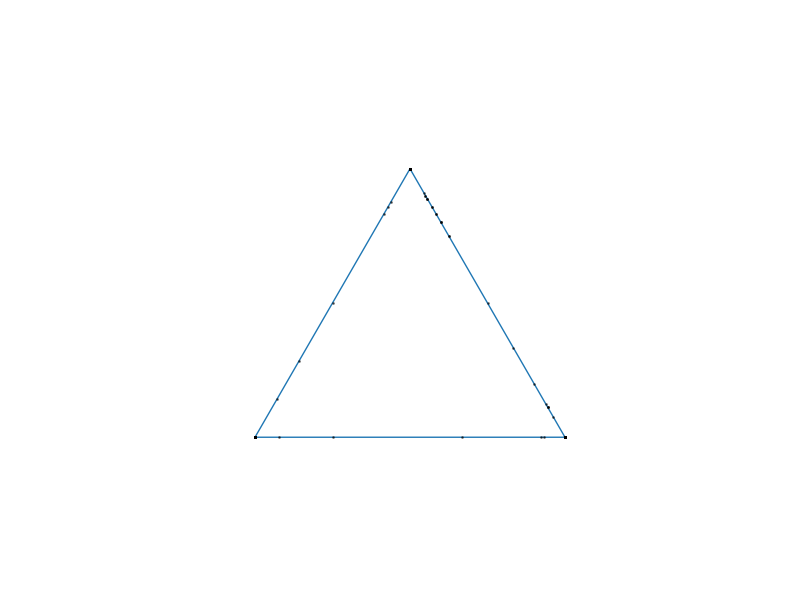}
  \end{subfigure}
  \begin{subfigure}[b]{0.32\textwidth}
  \centering
  \includegraphics[height=3.5cm,trim={4cm 4cm 4cm 4cm},clip]{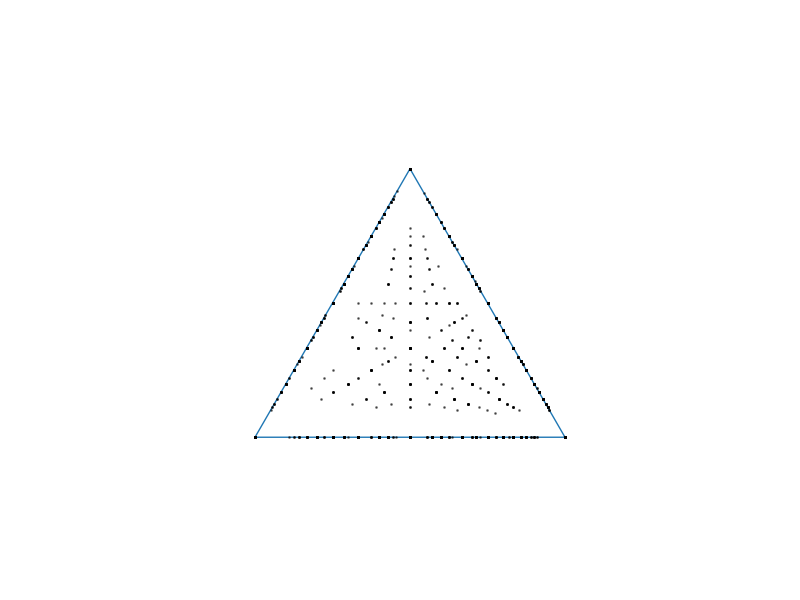}
  \end{subfigure}
  
  \caption{\label{fig:topic} Scatterplot of samples from the topic posterior $\mathbb{E}_{P_x}\Pgen(z|x)$. Left: real data. Middle: autoregressive LM generations. Right: non-autoregressive LM generations. Real data and autoregressive LM generations contain much less code-mixing compared to non-autoregressive LM generations.}
  \label{fig:pq}
\end{figure*}
\begin{table*}[!htp]
    \centering
    \begin{tabular}{@{}ll@{}}
    \toprule
    Topic 1 (German) &  die der und in zu den von für dass ist wir des nicht auf das eine werden es im auch\\
  Topic 2 (Spanish) & de la que en y el a los las del se una para un por no con es al sobre\\
  Topic 3 (French) & de la et des à les le que en ’ du dans nous pour qui une un est au pas     \\
\bottomrule
    \end{tabular}
    \caption{Learned topics largely correspond to languages. The top 20 words per topic are shown.}
    \label{tab:topics}
\end{table*}

To criticize the existence of code-mixing, we need a model that can model the mixing of languages of a document. LDA \citep{blei2003latent} is suitable for this purpose, as each language is analogous to a topic, and the document-topic latent variable parameterizes how topics (languages) are mixed in a document. 

In LDA, each document is associated with a topic coefficient latent variable $z \in \Delta(N)$, where $N=\{1,2,\ldots, M\}$ is a set of $M$ topics and $\Delta(N)$ is a probability simplex over these topics such that $z$ can be used to parameterize a categorical distribution.
The prior over $z$ is modeled using a Dirichlet distribution with parameters $\alpha$:
$$
P(z) \triangleq \text{Dirichlet}(z;\alpha).
$$

The document-topic coefficient latent variable $z$ defines a categorical distribution over the topics $t_n$ for each word $x_n$ in the document, and each topic in turn induces a categorical distribution over word types $P(x_n|t_n)=\phi_{t_n, x_n}$, where $\phi_{ij}$ parameterizes the probability of observing a word type $j$ conditioned on a topic $i$, so the joint probability of topics and words for a document with $N$ words is:
$$
P(x, t | z; \phi) = \prod_{n=1}^Nz_{t_n}\phi_{t_n, x_n}.
$$

Since we are only interested in criticizing the document-topic coefficient latent variable $z$, we marginalize out the topic assignments of each word. Assuming there are $M$ topics, the marginal distribution is:
$$
P(x|z;\phi) = \prod_{n=1}^N \sum_{i=1}^{M} z_i \phi_{i,x_n}.
$$

To fit this generative process on data, we set $M=3$ (since there are three target languages). We treat $\phi$ as a latent variable with prior $\text{Dirichlet}(z;\alpha)$, and then use collapsed Gibbs sampling to sample topic assignments $t$ from $P(t|x)$ (both $z$ and $\phi$ are collapsed). $\beta$ is fixed at 0.01, and $\alpha$ is optimized every 100 iterations with initial value $\frac{1}{M}$, and we use the MAP of $P(\phi|t,x)$ as a point estimate of $\phi^*$.\footnote{We use MALLET 2.0.8 \citep{mccallum2002mallet} to process data and learn the topic model.} For posterior inference, we use a two-stage sampling approach: Since $P(z, t | x; \phi^*) = P(t|x; \phi^*)P(z|t,x;\phi^*)$, we again apply collapsed Gibbs sampling to sample from $P(t|x; \phi^*)$ first, and then sampling from $P(z|t,x;\phi^*)$ is trivial since $P(z|t,x;\phi^*)=P(z|t; \phi^*) = \text{Dirichlet}(z;\alpha')$ where $\alpha_i'=\alpha_i + \sum_{n=1}^N \mathbbm{1}{ [t_n=i ]}$.

We evaluate two probabilistic formulations of transformer LMs in terms of code-mixing.  The first model is an autoregressive LM which assumes that each word depends on all previous words, and the second model is a non-autoregressive LM which assumes that different words are generated independent of each other \citep{gu2018non}. We train both LMs on the same English-German/Spanish/French dataset as in \citet{zhou2019understanding}.\footnote{We randomly split 80\% for training, 10\% for validation, and 10\% for testing, and the split might be different from \citet{zhou2019understanding}.} 

\paragraph{Model Settings} For both autoregressive and non-autoregressive LMs we use a transformer with 6 layers, 8 attention heads, model dimension 512, hidden dimension 2048.\footnote{We use the transformer\_wmt\_en\_de implementation in fairseq.} The autoregressive LM has 64.80M parameters, and the non-autoregressive LM has 65.98M parameters. Training takes about 16 hours on a single Nvidia A100 GPU.




\paragraph{Results} \Cref{tab:topics} shows the learned topics, which largely correspond to the three underlying languages. \Cref{fig:topic} visualizes samples from the posterior $P(z|x)$. We can see that for both the ground truth data and the autoregressive LM generations, the posterior is concentrated at the corners (hence it appears that there are fewer points), indicating that each translation contains mostly the same topic (underlying language). On the other hand, the posterior for non-autoregressive LM generations is dispersed, indicating that it's unable to fully commit to a single topic during generation due to the strong independence assumption. This result is the same as \citet{zhou2019understanding} without relying on external lexicons.

\section{Details of ``A Surprising Text Generation Failure''\label{sec:appendix_why_eval_latent}}
\paragraph{Data} We set $M$ to 50, $|\mathcal{Z}|$ to 256, $\mathcal{V}$ to the set of upper- and lower- case letters plus a special end-of-sequence symbol (so $|\mathcal{V}|=53$. We uniformly sample 10k distinct subsequences of tokens $x_1^m,\ldots,x_{N_m}^m$ by first sampling uniformly a length between 4 and 11, and then each token is drawn uniformly from the set of letters (except for the last token $x_{N_m}^m$, which is always end-of-sequence). For each subsequence of tokens, we sample uniformly from $\mathcal{Z}$ and only allow emissions from the sampled state to this subsequence (such that the posterior $P(z_m | x_1^m, \ldots, x_{N_m}^m)$ is a delta distribution). The entries in the transition matrix $\Pgen(z_m | z_{m-1})$ are initialized with a normal distribution, divided by temperature 0.5, and then normalized using softmax. The entries in the emission matrix $\Pgen(x_1^m,\ldots,x_{N_m}^m|z_m)$ are initialized with a normal distribution and divided by temperature 0.3. Then we mask out emissions not allowed and normalize the matrix using softmax. 

\paragraph{Posterior Inference} Given a sequence $x$, the goal of posterior inference is to infer $P(z|x)$. This can be done in two steps: first, we segment $x$ into subsequences (each subsequence corresponds to one hidden state). This segmentation is deterministic due to the end-of-sequence tokens. Next, we map each subsequence to its hidden state $z_m$ by a simple lookup operation because we masked the emission matric to only allow one hidden state per subsequence. Therefore, $P(z|x)$ is a delta distribution.

\paragraph{Model} The HSMM LM has 800 states. It is parameterized with the logits of its transition matrix $P(z_m | z_{m-1})$, its length emission matrix $P(N_m|z_m)$, and its emission matrix $P(x_1^m, \ldots, x_{N_m}^m|z_m, N_m)$. $N_m$ ranges from 1 to 11. To parameterize the emission matrix, we take the 250k most common n-grams in the training dataset for n from 1 to 11.\footnote{We cannot use the ground truth 10k valid subsequences of tokens since that would give HSMM LM an unfair advantage over the transformer LM.} It has 1.61B parameters due to the large number of possible emissions (the true data distribution $\Pdata$ only has 2.63M parameters). The transformer LM has 6 layers, 4 attention heads, model dimension 512, and hidden dimension 1024.\footnote{We use the transformer\_iwslt\_de\_en implementation in fairseq \citep{ott-etal-2019-fairseq}.} It has 18.94M parameters.

\paragraph{Optimization} We optimize HSMM using stochastic gradient descent (SGD) on the log marginal likelihood $\log P(x)$.\footnote{While HSMMs are usually optimized using the EM algorithm, we used SGD to be more comparable to transformers.} To marginalize out $z_m$ and $N_m$, we use PyTorch-Struct \citep{rush-2020-torch}. The model parameters are initialized with Xavier \citep{glorot2010understanding}. We use a batch size of 8 and train the model for 10 epochs with the Adam optimizer \citep{kingma2014adam} on an Nvidia A100 GPU. The learning rate is initialized to 3e-1 and halved when the validation log marginal likelihood does not improve for 240 steps, with a minimal learning rate 3e-4. We found it necessary to pretrain the emission matrix $P(x_1^m, \ldots, x_{N_m}^m|z_m, N_m)$ for one epoch using a learning rate of 1e-1 while fixing other parameters to avoid the under-utilization of states. Pretraining takes about a day and training takes about a week, due to the large number of parameters and the small batch size that we can afford. The transformer LM is optimized with Adam as well, but with a batch size of 4096 tokens, 4k learning rate warmup steps to maximum learning rate 5e-4. It is optimized to 120k steps in total (about 19 epochs), following fairseq's default setting for conditional language modeling on IWSLT14 De-En.\footnote{\url{https://github.com/pytorch/fairseq/blob/5e343f5f23b4a90cca2beec416b87d4dd7a4264f/examples/translation/README.md\#iwslt14-german-to-english-transformer}} Training the transformer LM takes about 4 hours on an Nvidia A100 GPU.

\begin{table*}[!htp]
    \centering
    \begin{tabular}{@{}l@{}rrrcccc@{}}
    \toprule
        Dataset & \#Train &  \#Val & \#Test & \#Sect Types & Med \#Sect & Med Sect Len & Max Sect Len  \\
        \midrule
        \textsc{PubMed} &  32.35k & 1.80k & 1.84k & 27 & 4 & 518 & 1986 \\
        \textsc{ArXiv} & 4.91k & 0.17k & 0.15k & 50 & 4 & 787.5 & 1965 \\
        \textsc{Wiki} & 111.40k & 13.97k & 13.98k & 96 & 4 & 122 & 1999 \\
        \textsc{Wiki-Short} &69.30k & 8.56k & 8.68k & 96 & 4 & 101 & 500\\
        \bottomrule
    \end{tabular}
    \caption{\label{tab:data}Data Statistics. Section length is measured using the GPT-2 tokenizer \citep{radford2019language,wolf-etal-2020-transformers}. Section statistics are based on the validation set. More details on data processing can be found in \Cref{sec:appendix_section}.}
\end{table*}

\section{Details of ``Critiquing Discourse Coherence''\label{sec:appendix_section}}
\paragraph{Data - \textsc{PubMed} and \textsc{ArXiv}} The \textsc{PubMed} and \textsc{ArXiv} datasets in \Cref{sec:dc} are based on the datasets of \citet{cohan-etal-2018-discourse}, where each article consists of a list of sections with section titles.\footnote{They use an Apache-2.0 license.} We process the dataset in a few steps: First, we standardize the section titles by lemmatizing each word in the section title,\footnote{We use the lemmatizer of NLTK 3.6.7 \citep{bird-loper-2004-nltk}.} removing any numbers, and mapping each word to a standard spelling (e.g., ``acknowledgement'' is mapped to ``acknowledgment''). Next, we remove from each article ``see also'', ``external link'', ``reference'', ``further reading'', ``note'', and ``source'' sections. Then we filter articles with fewer than 3 remaining sections, or with sections of more than 2k tokens or fewer than 30 tokens (the number of tokens is counted according to the GPT-2 tokenizer). Finally, we remove articles containing infrequent section titles, where the threshold is 500 for \textsc{PubMed} and 200 for \textsc{ArXiv} (all counted on the training dataset).

\paragraph{Data - \textsc{Wiki}} We download the English Wikipedia dumped on Dec 1, 2021.\footnote{\url{ https://dumps.wikimedia.org/enwiki/20211201/}} We then use a Python package mwparserfromhell\footnote{\url{https://github.com/earwig/mwparserfromhell} (version 0.7.dev0)} to extract top-level sections from each article. We ignore redirect pages, disambiguation pages, links, files, and images, and strip away code. We also ignore articles about years. Then we process the dataset in the same way as how we processed \textsc{PubMed} and \textsc{ArXiv}, except that we remove articles with fewer than 4 sections (since we always count the introductory paragraph of each article in Wikipedia as an ``abstract'' section), and that we removed infrequent section titles that appear fewer than 4k times in the training data.

\paragraph{Dataset Statistics} The statistics of all three datasets can be found at \Cref{tab:data}.

\begin{table}[!t]
    \centering
    \begin{tabular}{@{}llc@{}}
    \toprule
       $z_{m-1}$ & $z_{m}$ & Frequency (\%) \\
        \midrule
         W/O Title\\
          career & career & 0.26\\
          abstract & life and career & 0.24\\
          reception & reception & 0.14\\
          abstract & playing career & 0.14\\
          plot & plot & 0.14\\
          total failures & -& 7.54\\
         \midrule
         W/ Title\\
          abstract & playing career & 0.16\\
          abstract & life and career & 0.14\\
          career & career & 0.13\\
          abstract & production & 0.11\\
          total failures & - & 5.46\\
         \bottomrule
    \end{tabular}
    \caption{The top 5 section transition errors on \textsc{Wiki}. Frequency is the frequency of observing the specific transition error across all transitions in the (generated) dataset. A transition is deemed an error if $\Pgen(z_{m}|z_{m-1}) < 0.01$. Here we use the better-performing GPT-2-based LMs (LM$_1$).}
    \label{tab:unlikely}
\end{table}

\begin{table*}[!htp]
    \centering
    \begin{tabular}{@{}lp{12cm}@{}}
    \toprule
        Inferred Section Title & Section Text  \\
        \midrule
        abstract & Toledo Township is a township in Dauphin County, Pennsylvania, United States. As of the 2010 census, the township population was 1,655. It is part of the Dutchess/Berwick Micropolitan Statistical Area. \newline Toledo Township was organized in 1867, and named for the United States senator, Judge John Toledo.  \\
        geography & Toledo Township is in northern Dutchess County, bordered by the city of Altoona to the north, the borough of Ritzel to the southeast, and Tuskegee Township and the unincorporated hamlet of Westmoreland Township to the south. According to the United States Census Bureau, the township has a total area of, of which is land and, or 0.62\%, is water. It is bordered on the south by the Tullahonas River, on the west by the Delaware Channel, on the south by the Mohawk River and on the west by Tullahonas Creek, whose tributaries are the Westmoreland and Trenton rivers. Pennsylvania Route 11, which runs between Routes 11 and N, crosses the township via the Tuskegee River ... \\
        {\color{red}demographic} & As of the census of 2000, there were 1,638 people, 809 households, and 595 families residing in the township. The population density was 787.1 people per square mile (285.2/km2). There were 944 housing units at an average density of 331.2 per square mile (126.5/km2). The racial makeup of the township was 95.07\% White, 1.81\% African American, 0.46\% Native American, 0.36\% Asian, 0.06\% Pacific Islander, 0.42\% from other races, and 1.06\% from two or more races. Hispanic or Latino of any race were 1.13\% of the population. \newline There were 809 households, out of which 32.4\% had children under the age of 18 living with them, 49.0\% were married couples living together, 11.1\% had a female householder with no husband present, and 30.0\% were non-families. 26.5\% of all households were made up of individuals, and 12.9\% had someone living alone who was 65 years of age or older ...\\
        {\color{red}demographic} & Census 2010 \newline As of the 2010 United States Census, there were 1,655 people, 613 households, and 585 families residing in the township. The population density was 847.8 people per square mile (287.1/km2). There were 640 housing units at an average density of 296.1 per square mile (110.2/km2). The racial makeup of the township was 95.17\% White, 1.81\% African American, 0.41\% Native American, 0.12\% Asian, 1.00\% from other races, and 0.49\% from two or more races. Hispanic or Latino of any race were 2.67\% of the population. \newline There were 613 households, out of which 33.8\% had children under the age of 18 living with them, 56.9\% were married couples living together, 12.7\% had a female householder with       no husband present, and 29.7\% were non-families. 24.6\% of all households were made up of individuals, and 13.1\% had someone living alone who was 65 years of age or older. \\
        notable people & Joseph R. Clements (May 17, 1911 – June 23, 1998) Mayor of Mount Pleasant, South Carolina \newline Jefferson Daugherty (born 1935 in Chatham) U.S. Senator, United States House of Representatives ...\\
         \bottomrule
    \end{tabular}
    \caption{An example section-level repetition error (LM$_1$~W/O Title on \textsc{Wiki}). The most common section type after ``demographic'' is ``education''. The structures of the repeated sections are similar yet the facts are different.}
    \label{tab:example_repetition}
\end{table*}

\paragraph{Posterior Inference} We finetune a BERT classifier \citep{devlin-etal-2019-bert} to estimate $P(z_m|x_m)$ using the Adam optimizer. We use a batch size of 32, learning rate of 2e-5, and finetune for 3 epochs. The validation accuracies are 89.48\%, 72.52\%, and 88.15\% on \textsc{PubMed}, \textsc{ArXiv}, and \textsc{Wiki} respectively. Finetuning takes up to a few hours on a single Nvidia A100 GPU.

\paragraph{Language Models} We use the base version of GPT-2\footnote{\url{https://huggingface.co/gpt2}} (LM$_1$, which has about 117M parameters) and the 125M version of GPT-Neo (LM$_2$)\footnote{\url{https://huggingface.co/EleutherAI/gpt-neo-125M}}. We use Adam to finetune all LMs.\footnote{We use the training script from \url{https://github.com/huggingface/transformers/blob/master/examples/legacy/run_language_modeling.py}.} Since both LM$_1$ and LM$_2$ have a limited context window size, we use truncated BPTT \citep{puskorius1994truncated} with context window size set to the maximum value possible (1024 for LM$_1$ and 2048 for LM$_2$). At generation time, we generate one token at a time and truncate the context to fit within the context window. We use a special symbol \verb!<endoftext>! to mark article boundaries. For optimization we use a batch size of 8 (for GPT-Neo-based LMs we use a batch size of 4 but update parameters every two steps), a learning rate of 5e-5 (we did an initial learning rate search from \{5e-6, 5e-5, 5e-4, 5e-3\} on \textsc{PubMed} and found 5e-5 to perform the best), and train the model for 20 epochs. Model checkpoints with the best validation loss (the lowest validation PPL) are used for the final evaluation. Training takes up to 24 hours on a single Nvidia A100 GPU.

\paragraph{Error Analysis} \Cref{tab:unlikely} presents the most common section transition errors across all section types for different settings (W/ Title and W/O Title). %
We notice again that a very common transition error is to generate a section repeatedly. However, even in the ground truth data, there are repeated sections, such as ``career $\to$ career'' (appearing 0.08\%), which is due to the misclassification by the BERT classifier used for the inference network.\footnote{In the training data there also exist some rare repetitions, such as \url{https://web.archive.org/web/20220307192058/https://en.wikipedia.org/wiki/Leanna_Brown}.}

\paragraph{Repetition Errors} Repetition is a common type of error found by model criticism (see \Cref{tab:unlikely}). For example, on the \textsc{Wiki} dataset, repetition errors account for 25.93\% of all errors (using the same criterion as in \Cref{tab:unlikely}) on LM$_1$~W/O Title and 17.89\% on LM$_1$~W/ Title. While previous works have shown that neural language models tend to repeat at the level of phrases \citep{holtzman2019curious} and sentences \citep{welleck2019neural}, our work found that the repetition might even happen at a higher level, as shown in the qualitative example in \Cref{tab:example_repetition}.



\begin{figure*}[t]
  \centering
  \includegraphics[width=\textwidth,trim={0cm 0cm 0cm 0cm},clip]{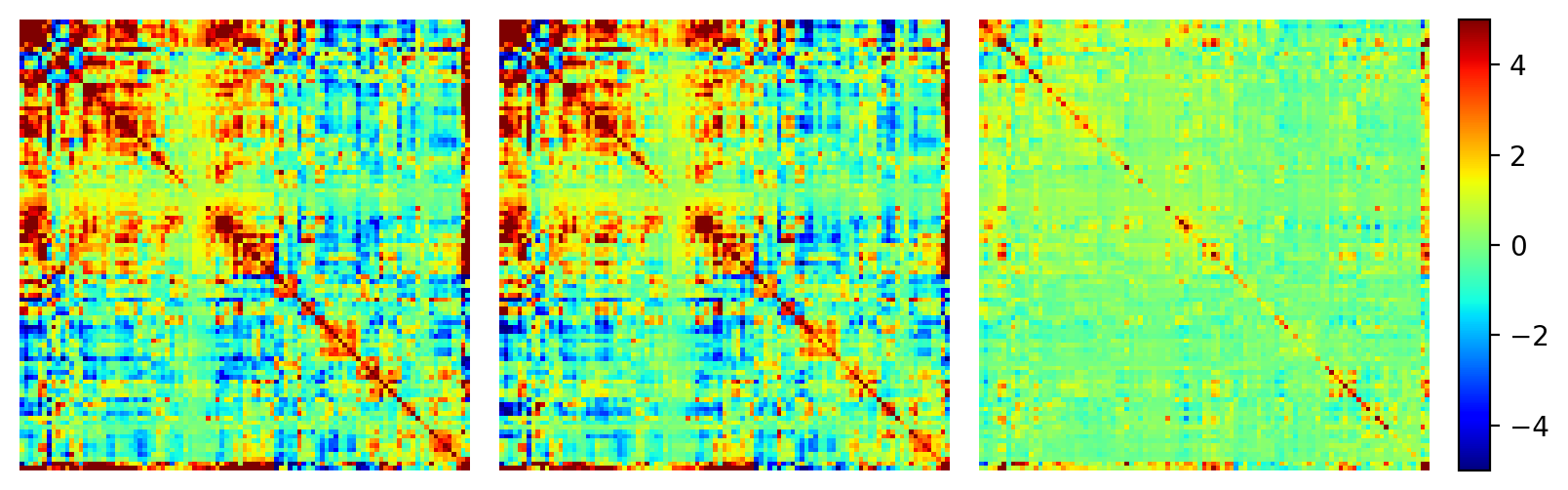}
  \caption{\label{fig:cov_pubmed} Topic covariance matrix for the induced $z$ (on \textsc{PubMed}). Left: Test set ($\Pdata$). Middle: LM$_1$ generations ($\Pmodel$). Right: generations of LM$_1$ trained on \textsc{ArXiv} as a visual baseline.}
\end{figure*}
\begin{figure*}[!htp]
  \centering
  \includegraphics[width=\textwidth,trim={0cm 0cm 0cm 0cm},clip]{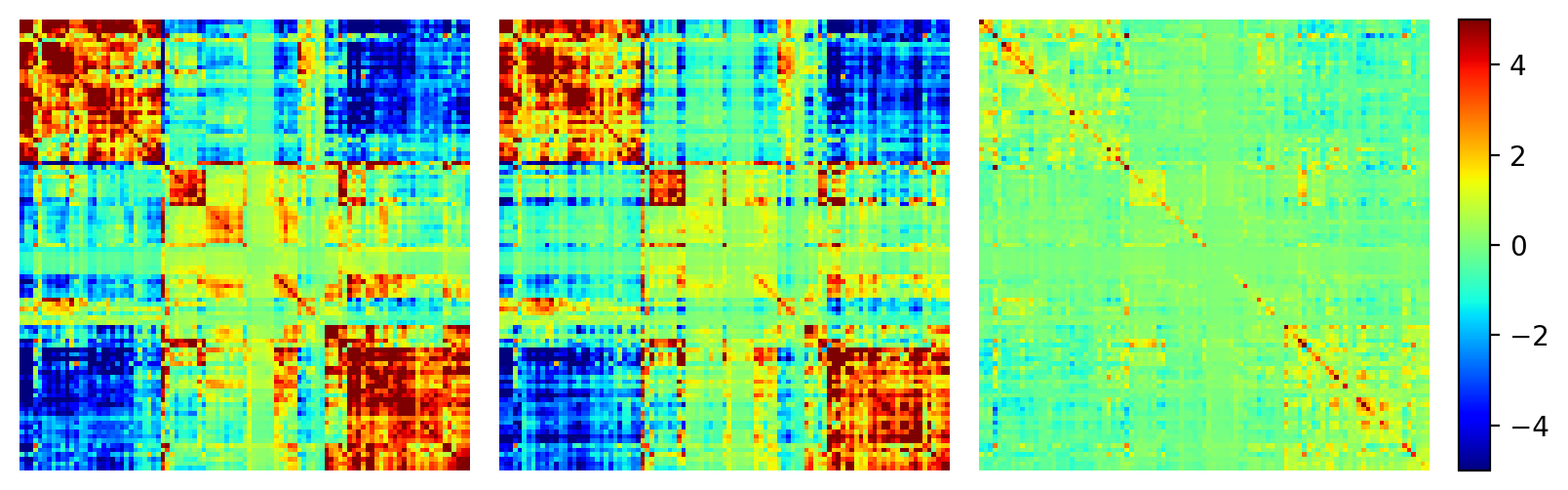}
  \caption{\label{fig:cov_arxiv} Topic covariance matrix for the induced $z$ (on \textsc{ArXiv}). Left: Test set ($\Pdata$). Middle:  LM$_1$ generations ($\Pmodel$). Right: generations of LM$_1$ trained on \textsc{PubMed} as a visual baseline.}
\end{figure*}

\begin{table*}[!htp]
    \centering
    \begin{tabular}{@{}lp{14cm}@{}}
    \toprule
        $T(x)$ & Text \\
        \midrule
        280.88 & ...  to the best of our knowledge, this result, together with previous results, supports the conclusion that there is no difference in the spin concentration between bp and qp versions of the hamiltonian in the quenched version of the hamiltonian in the quenched version of the hamiltonian in the qp version of the hamiltonian in the quenched version of the hamiltonian in the quenched version of the hamiltonian in the quenched version of the hamiltonian in the quenched ver     sion of the hamiltonian in the quenched version of the hamiltonian in the quenched version of the hamiltonian in the quenched version of the hamiltonian in the quenched version ...\\ 
        224.74 & ... if we now use the electrostatic potential of the electrostatic potential of the electrostatic potential of the electrostatic potential of the electrostatic potential of the electrostatic potential of the electrostatic potential of the electrostatic potential of the electrostatic potential of the electrostatic potential of the electrostatic potential of the electrostatic potential of the electrostatic potential of the electrostatic potential ...\\
        \bottomrule
    \end{tabular}
    \caption{Top 2 outliers identified by $T(x)$ on the generations from LM$_1$ finetuned on the \textsc{ArXiv} dataset (W/O Title). The average $T(x)$ (Latent NLL) is 161.40.}
    \label{tab:outliers}
\end{table*}
\section{Details of ``Critiquing Topic Correlations'' \label{sec:appendix_topic}}
\paragraph{Data} We use the same datasets as in \Cref{sec:dc}. For topic modeling, we remove word types that appear in more than 50\% of training documents, and we also remove \LaTeX commands such as \verb!\xmath!.

\paragraph{Topic Model Training} We use $M=100$ topics. To learn the topic model, we use variational EM to optimize the ELBO with the default training settings in David Blei's CTM implementation.\footnote{\url{http://www.cs.columbia.edu/~blei/ctm-c/}} At inference, we also use variational inference (without the M step), also with the default inference settings in David Blei's CTM implementation. Training takes up to a few days using a single Intel Xeon Platinum 8358 CPU.

\paragraph{Outlier Detection} While \Cref{sec:topic} has shown that in aggregate the Latent NLL of the LM$_1$ generations is close to that of real data, we can identify outliers by finding $x$ for which $T(x)=-\mathbb{E}_{z\sim \Pgen(z|x)} \log \Pgen(z)$ is high. We find that those outliers are usually pathological cases that result in a very different distribution of topics, as shown in \Cref{tab:outliers}.

\paragraph{More Visualizations} \Cref{sec:topic} visualized the covariance matrices on \textsc{Wiki}. We also plot the covariance matrices on \textsc{PubMed} and \textsc{ArXiv} in \Cref{fig:cov_pubmed} and \Cref{fig:cov_arxiv}. Note that we use hierarchical clustering of the covariance matrix on the test set to reorder topics, and we clamp the values in the covariance matrix to be in the range of [-5, 5] for plotting.

\section{Details of ``Critiquing Coreference Chains'' \label{sec:appendix_coref}}
\paragraph{Data} All experiments in this section use a subset of the \textsc{Wiki} dataset: we apply a simple filter to only consider articles about films, by matching the first section of the article with the regular expression \verb!.*is a.*film.*!.

\paragraph{Coreference Resolution} We use an off-the-shelf neural coreference resolution system neuralcoref\footnote{\url{https://github.com/huggingface/neuralcoref/tree/60338df6f9b0a44a6728b442193b7c66653b0731}} to infer $\mathbf{z}$ given an article. We limit our studies to only consider person entities.

\paragraph{Gender Assignment} To avoid the open vocabulary problem of proper nouns and also due to the fact that personal pronouns are usually gendered in English, we replace proper nouns with their genders (Male/Female/Plural/None of the above). In order to identify genders of proper nouns, we use the majority voting of the genders of the pronouns that corefer with them (for example, ``she'' corresponds to female, ``he'' corresponds to male, and ``they'' corresponds to plural). If there are no gendered pronouns that corefer with the given proper noun, we assign ``None of the above'' as the gender. The caption of \Cref{fig:coref_qual3} presents an example of the gender assignment procedure.

\paragraph{Language Models} We use LM$_1$ (W/O Title) trained on \textsc{Wiki}. We apply the same filtering process as applied to real data to only consider generations about films.

\begin{table}[!htp]
\small
    \centering
    \begin{tabular}{@{}l@{}l@{}l@{}l@{}l@{}c@{ }c@{ }c@{}}
    \toprule
       $z_{1}$ & $z_2$ & $z_3$ & $z_4$ & $z_5$ & $\log P_c$ & $z_5^*$ & $\log P_c(z_5^*|z_{<5})$\\
        \midrule\relax
[M]$_0$  & [he]$_0$  & [M]$_0$  & . & [They]$_1$  & -13.37 & [M]$_0$  & -1.22\\\relax
. & [M]$_0$  & [M]$_1$  & [M]$_2$  & [she]$_2$  & -14.72 & [his]$_2$  & -2.72\\\relax
  & . & [M]$_0$  & . & [her]$_1$  & -11.29 & [He]$_0$  & -1.67\\\relax
[N]$_0$  & . & [M]$_1$  & [N]$_0$  & [she]$_2$  & -8.05 & [his]$_1$  & -2.15\\\relax
[F]$_0$  & [her]$_0$  & [he]$_1$  & [M]$_2$  & [he]$_2$  & -7.03 & [his]$_1$  & -2.22\\\relax
[he]$_0$  & [her]$_1$  & [she]$_1$  & . & [He]$_2$  & -8.14 & [M]$_0$  & -1.87\\\relax
. & . & [M]$_0$  & [he]$_1$  & [she]$_0$  & -14.18 & [his]$_1$  & -1.48\\\relax
. & [N]$_0$  & . & [he]$_1$  & [her]$_2$  & -10.82 & [his]$_1$  & -1.48\\\relax
[M]$_0$  & . & [her]$_1$  & . & [He]$_2$  & -8.74 & [M]$_0$  & -2.20\\\relax
[F]$_0$  & [she]$_0$  & . & [F]$_0$  & [him]$_1$  & -8.18 & [her]$_0$  & -1.28\\\relax
[F]$_0$  & [M]$_1$  & . & [F]$_2$  & [him]$_3$  & -8.56 & [her]$_2$  & -1.41\\\relax
[F]$_0$  & [F]$_1$  & [her]$_1$  & [her]$_1$  & [he]$_1$  & -10.25 & [her]$_1$  & -1.72\\\relax
[F]$_0$  & [her]$_0$  & [her]$_0$  & [he]$_0$  & [P]$_1$  & -7.28 & [her]$_0$  & -1.23\\
         \bottomrule
    \end{tabular}
    \caption{Unlikely $z_5|z_{<5}$ and the corresponding $\log P_c(z_5|z_{<5})$ in LM$_1$ generations according to the learned critic ($\log P_c(z_5|z_{<5}) < -7$). To get a better sense of what is considered likely by the critic, we also showed $z_5^*=\arg\max_{z_5} P_c(z_5|z_{<5})$ as well as $\log P_c(z_5^*|z_{<5})$. M (Male), F (Female), P (Plural), and N (None of the above). Blank: padding.}
    \label{tab:result_critic}
\end{table}

\paragraph{Critic} We use a 5-gram language model with Kneser-Ney smoothing \citep{ney1994structuring} to fit the critic distribution, where we used $\frac{\text{\#unique (n-1)-grams}}{\text{\#unique (n-1)-grams}+2\text{\#unique n-grams}}$ as the discount factor \citep{stolcke2002srilm}. 

What does this critic learn? \Cref{tab:result_critic} shows a random subset of unlikely coreference chain n-grams generated by LM$_1$ according to the critic. We can see that the learned critic makes sense intuitively. For example, in the first row, [They]$_1$ is created even though the previous context only contains a single entity;\footnote{That being said, it is possible that outside this context window there are other entities that makes using ``They'' possible.} in the second row, ``she'' is used to refer to a male; in the third row, ``her'' doesn't have any antecedent.\footnote{Since this 5-gram starts with padding, there is nothing to the left of the context window.}

\begin{table}[!t]
\small
    \centering
    \begin{tabular}{@{}lllllcc@{}}
    \toprule
       $z_{1}$ & $z_2$ & $z_3$ & $z_4$ & $z_5$ & $P_{\text{data}}$ & $P_{\text{LM}}$\\
        \midrule
        . & . & [M]$_0$ & [F]$_1$ & [M]$_0$ & 0.01 & 0.06 \\\relax
        [M]$_0$ & . & [N]$_1$ & [M]$_0$ & [M]$_2$ & 0.04 & 0.17 \\\relax
        [M]$_0$ & . & [N]$_1$ & [M]$_0$ & [M]$_0$ & 0.04 & 0.17 \\
        . & . & [M]$_0$ & [he]$_0$ & [M]$_0$ & 0.01 & 0.04 \\
        . & [He]$_0$ & [his]$_0$ & . & [His]$_0$ &  0.02 & 0.06 \\
         \bottomrule
    \end{tabular}
    \caption{The top 5 coreference chain n-grams with the largest log probability differences between LM generations and real data ($\log P_{\text{LM}}(z_5|z_{<5}) - \log P_{\text{data}}(z_5|z_{<5})$). We only consider n-grams that appear more than (including) 5 times in both test set and LM generations. M (Male), F (Female), and N (None of the above).}
    \label{tab:result_coref}
\end{table}

\begin{table}[!htp]
\small
    \centering
    \begin{tabular}{@{}lllllcc@{}}
    \toprule
       $z_{1}$ & $z_2$ & $z_3$ & $z_4$ & $z_5$ & $P_{\text{data}}$ & $P_{\text{LM}}$\\
        \midrule
        . & [N]$_0$ & [F]$_1$ & . & [M]$_2$ & 0.17 & 0.05 \\\relax
        [him]$_0$ & [she]$_1$ & [him]$_0$ & . & . & 0.48 & 0.14 \\\relax
        [M]$_0$ & [him]$_0$ & [he]$_0$ & . & . & 0.36 & 0.11 \\\relax
        [N]$_0$ & [N]$_0$ & [N]$_0$ & . & [N]$_0$ &  0.18 & 1.16 \\
         &  &  & . & [P]$_0$ & 0.01 & 0.00 \\
         \bottomrule
    \end{tabular}
    \caption{The top 5 coreference chain n-grams with the \emph{lowest} log probability differences between LM generations and real data ($\log P_{\text{LM}}(z_5|z_{<5}) - \log P_{\text{data}}(z_5|z_{<5})$). We only consider n-grams that appear more than (including) 5 times in both test set and LM generations. M (Male), F (Female), P (Plural), and N (None of the above). Blank: padding.}
    \label{tab:result_coref_bottom}
\end{table}


\paragraph{More Results} \Cref{tab:result_coref} shows the coreference chains that occur more frequently in LM generations than in real data (we again used 5-gram LM with Kneser-Ney smoothing to estimate the probabilities). We can see that some of these are implausible similar to the observation in the main paper: for example, in the second to last row a proper noun [Male]$_0$ is used after a pronoun [he]$_0$ is used to refer to the same entity in the sentence.  In \Cref{tab:result_coref_bottom} we show the other direction: the coreference chains that occur more frequently in real data than LM generations. We can see that while this also shows the places where the coreference distributions do not match, the coreference structures here are not unlikely, since they appear frequently in real data.



\begin{figure}[!t]
    \begin{framed}
     \textbf{Tokenized Text}
    
    ... . After the marriage , \mybox[fill=green!20]{[Jyothish]$_0$} finds out that Raja \mybox[fill=blue!20]{[Rao]$_1$} had raped \mybox[fill=yellow!20]{[Sridevi]$_2$} . and \mybox[fill=blue!20]{[he]$_1$} also tells \mybox[fill=blue!20]{[him]$_1$} that Raja \myboxtwo[fill=blue!20]{[Rao]$_1$} is \mybox[fill=blue!20]{[his]$_1$} father , so \mybox[fill=blue!20]{[he]$_1$} tries to kill them both ... \\
    
    \textbf{Coreference Chains} $\mathbf{z}$
    
     . \mybox[fill=green!20]{[Female]$_0$} \mybox[fill=blue!20]{[Male]$_1$} \mybox[fill=yellow!20]{[None]$_2$} . \mybox[fill=blue!20]{[he]$_1$} \mybox[fill=blue!20]{[him]$_1$} \myboxtwo[fill=blue!20]{[Male]$_1$} \mybox[fill=blue!20]{[his]$_1$} \mybox[fill=blue!20]{[he]$_1$} \\
    
    \textbf{5-gram critic} $\mathbf{P_c}$
    \begin{align*}
        &P_c(\ \myboxtwo[fill=blue!20]{\text{[Male]$_1$}}\ | \text{previous entity mentions}) \\
        &\approx P_c(\ \myboxtwo[fill=blue!20]{\text{[Male]$_1$}}\ |\ \mybox[fill=yellow!20]{\text{[None]$_2$}} \text{.} \mybox[fill=blue!20]{\text{[he]$_1$}} \mybox[fill=blue!20]{\text{[him]$_1$}})\\
        & = \exp(-7.29)
    \end{align*}
    \end{framed}
    \caption{\label{fig:coref_qual1}A qualitative example where the critic correctly identifies an implausible coreference n-gram. The argmax at the circled position is \mybox[fill=blue!20]{[he]$_1$} with probability $\exp(-2.32)$. We only highlighted the root of each entity mention to avoid clutter.
    }
\end{figure}

\begin{figure}[!htp]
    \begin{framed}
     \textbf{Tokenized Text}
    
    ... . \mybox[fill=green!20]{[He]$_0$} realizes that \mybox[fill=green!20]{[he]$_0$} must put \mybox[fill=green!20]{[his]$_0$} life so that the \mybox[fill=blue!20]{[girl]$_1$} will know anything about it , only that \mybox[fill=blue!20]{[she]$_1$} wo n't because of \mybox[fill=green!20]{[himself]$_0$} ... \\
    
    \textbf{Coreference Chains} $\mathbf{z}$
    
     . \mybox[fill=green!20]{[He]$_0$} \mybox[fill=green!20]{[he]$_0$} \mybox[fill=green!20]{[his]$_0$} \mybox[fill=blue!20]{[girl]$_1$} \mybox[fill=blue!20]{[she]$_1$} \mybox[fill=green!20]{[himself]$_0$} \\
    
    \textbf{5-gram critic} $\mathbf{P_c}$
    \begin{align*}
        &P_c(\ \myboxtwo[fill=green!20]{\text{[himself]$_0$}}\ | \text{previous entity mentions}) \\
        &\approx P_c(\ \myboxtwo[fill=green!20]{\text{[himself]$_0$}}\ |\ \mybox[fill=green!20]{\text{[he]$_0$}} \mybox[fill=green!20]{\text{[his]$_0$}} \mybox[fill=blue!20]{\text{[girl]$_1$}} \mybox[fill=blue!20]{\text{[she]$_1$}})\\
        & = \exp(-9.00)
    \end{align*}
    \end{framed}
    \caption{\label{fig:coref_qual4}A qualitative example where the critic correctly identifies an implausible coreference n-gram. The argmax at the circled position is \mybox[fill=green!20]{[him]$_0$} with probability $\exp(-1.43)$. We only highlighted the root of each entity mention to avoid clutter.
    }
\end{figure}

\makeatletter
\setlength{\@fptop}{0pt}
\makeatother

\begin{figure}[t!]
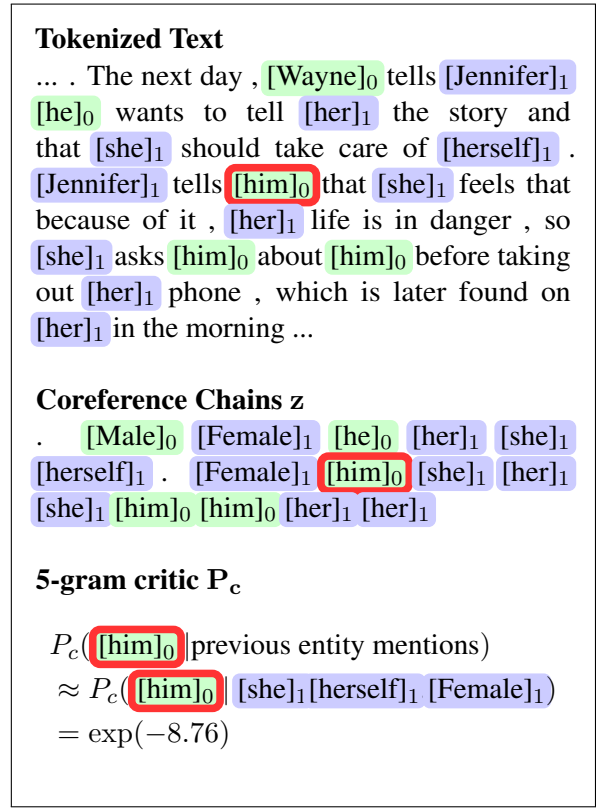

    \begin{framed}
     \textbf{Tokenized Text}
    
    ... . The next day , \mybox[fill=green!20]{[Wayne]$_0$} tells \mybox[fill=blue!20]{[Jennifer]$_1$} \mybox[fill=green!20]{[he]$_0$} wants to tell \mybox[fill=blue!20]{[her]$_1$} the story and that \mybox[fill=blue!20]{[she]$_1$} should take care of \mybox[fill=blue!20]{[herself]$_1$} . \mybox[fill=blue!20]{[Jennifer]$_1$} tells \myboxtwo[fill=green!20]{[him]$_0$} that \mybox[fill=blue!20]{[she]$_1$} feels that because of it , \mybox[fill=blue!20]{[her]$_1$} life is in danger , so \mybox[fill=blue!20]{[she]$_1$} asks \mybox[fill=green!20]{[him]$_0$} about \mybox[fill=green!20]{[him]$_0$} before taking out \mybox[fill=blue!20]{[her]$_1$} phone , which is later found on \mybox[fill=blue!20]{[her]$_1$} in the morning ... \\
    
    \textbf{Coreference Chains} $\mathbf{z}$
    
     . \mybox[fill=green!20]{[Male]$_0$} \mybox[fill=blue!20]{[Female]$_1$} \mybox[fill=green!20]{[he]$_0$} \mybox[fill=blue!20]{[her]$_1$} \mybox[fill=blue!20]{[she]$_1$} \mybox[fill=blue!20]{[herself]$_1$} . \mybox[fill=blue!20]{[Female]$_1$} \myboxtwo[fill=green!20]{[him]$_0$} \mybox[fill=blue!20]{[she]$_1$} \mybox[fill=blue!20]{[her]$_1$}  \mybox[fill=blue!20]{[she]$_1$} \mybox[fill=green!20]{[him]$_0$} \mybox[fill=green!20]{[him]$_0$} \mybox[fill=blue!20]{[her]$_1$} \mybox[fill=blue!20]{[her]$_1$}\\
    
    \textbf{5-gram critic} $\mathbf{P_c}$
    \begin{align*}
        &P_c(\ \myboxtwo[fill=green!20]{\text{[him]$_0$}}\ | \text{previous entity mentions}) \\
        &\approx P_c(\ \myboxtwo[fill=green!20]{\text{[him]$_0$}}\ |\ \mybox[fill=blue!20]{\text{[she]$_1$}} \mybox[fill=blue!20]{\text{[herself]$_1$}} \text{.} \mybox[fill=blue!20]{\text{[Female]$_1$}} )\\
        & = \exp(-8.76)
    \end{align*}
    \end{framed}
    \caption{\label{fig:coref_qual2}A qualitative example where the critic incorrectly identifies an implausible coreference n-gram, due to the limited context window not containing the antecedent of the pronoun. The argmax at the circled position is \mybox[fill=blue!20]{[her]$_1$} with probability $\exp(-1.49)$. We only highlighted the root of each entity mention to avoid clutter.
    }
\end{figure}

\begin{figure}[t!]
    \begin{framed}
     \textbf{Tokenized Text}
    
    ... . \mybox[fill=green!20]{[Jack]$_0$} 's girlfriend , the wealthy Baron von Brühl ( Peter Lorre ) , also steals the \mybox[fill=blue!20]{[girl]$_1$} . After an unpleasant and embarrassing incident in which \mybox[fill=blue!20]{[she]$_1$} is forced to drink a pager before going home , the \mybox[fill=yellow!20]{[Baron]$_2$} 's \mybox[fill=pink!20]{[henchwoman]$_3$} is caught and thrown on the balcony of \mybox[fill=pink!20]{[his]$_3$} inn , where \mybox[fill=blue!20]{[she]$_1$} is set upon . \mybox[fill=green!20]{[Jack]$_0$} rescues \mybox[fill=blue!20]{[her]$_1$} and takes \mybox[fill=blue!20]{[her]$_1$} to Austria to live with von \mybox[fill=brown!20]{[Brühl]$_4$} . Von \mybox[fill=brown!20]{[Brühl]$_4$} is now worried that Jenny ( Ann Sheridan ) is in love with \mybox[fill=green!20]{[Jack]$_0$} . After realizing that \mybox[fill=brown!20]{[she]$_4$} is already engaged to \mybox[fill=green!20]{[Jack]$_0$} , \myboxtwo[fill=brown!20]{[he]$_4$} persuades \mybox[fill=brown!20]{[her]$_4$} to go with \mybox[fill=brown!20]{[him]$_4$} to Austria as soon as \mybox[fill=gray!20]{[they]$_5$} can ... \\
    
    \textbf{Coreference Chains} $\mathbf{z}$
    
     . \mybox[fill=green!20]{[None]$_0$} \mybox[fill=blue!20]{[Female]$_1$} . \mybox[fill=blue!20]{[she]$_1$} \mybox[fill=yellow!20]{[Male]$_2$} \mybox[fill=pink!20]{[Male]$_3$} \mybox[fill=pink!20]{[his]$_3$} \mybox[fill=blue!20]{[she]$_1$} . \mybox[fill=green!20]{[None]$_0$} \mybox[fill=blue!20]{[her]$_1$} \mybox[fill=blue!20]{[her]$_1$} \mybox[fill=brown!20]{[Female]$_4$} . \mybox[fill=brown!20]{[Female]$_4$} \mybox[fill=green!20]{[None]$_0$} . \mybox[fill=brown!20]{[she]$_4$} \mybox[fill=green!20]{[None]$_0$} \myboxtwo[fill=brown!20]{[he]$_4$} \mybox[fill=brown!20]{[her]$_4$} \mybox[fill=brown!20]{[him]$_4$} \mybox[fill=gray!20]{[they]$_5$}\\
    
    \textbf{5-gram critic} $\mathbf{P_c}$
    \begin{align*}
        &P_c(\ \myboxtwo[fill=brown!20]{\text{[he]$_4$}}\ | \text{previous entity mentions}) \\
        &\approx P_c(\ \myboxtwo[fill=brown!20]{\text{[he]$_4$}}\ |\ \mybox[fill=brown!20]{\text{[Female]$_4$}} \mybox[fill=green!20]{\text{[None]$_0$}} \mybox[fill=brown!20]{\text{[she]$_4$}} \mybox[fill=green!20]{\text{[None]$_0$}})\\
        &=\exp(-11.68)
    \end{align*}
    \end{framed}
    \caption{\label{fig:coref_qual3}A qualitative example where coreference errors are intertwined with coreference resolution errors: the circled position is deemed implausible because it's using a male pronoun \mybox[fill=brown!20]{[he]$_4$} to refer to a female \mybox[fill=brown!20]{[Brühl]$_4$}. The argmax at the circled position is \mybox[fill=brown!20]{[she]$_4$} with probability $\exp(-1.80)$. This example also shows how gender assignments are made: since \mybox[fill=pink!20]{[henchwoman]$_3$} corefers with a male pronoun \mybox[fill=pink!20]{[his]$_3$}, it is labeled as a male. We only highlighted the root of each entity mention to avoid clutter.
    }
\end{figure}

\paragraph{Qualitative Examples} \Cref{fig:coref_qual1} and \Cref{fig:coref_qual4} show two examples where coreference abnormalities are successfully detected by the model. \Cref{fig:coref_qual2} shows an example where due to the limited context window size of the 5-gram critic, a pronoun is identified as unlikely due to its antecedent falling outside the context window even though it is appropriate. \Cref{fig:coref_qual3} shows an example where due to coreference resolution errors are intertwined with coreference errors. This type of errors would likely go away as more powerful coreference resolution systems are developed. 

\paragraph{Potential Improvements} By throwing away all the other words but the entity mentions, we lose much information about the sentence, even syntactic information such as the c-command structures \citep{chomsky1993lectures}. By augmenting the entity mentions with syntactic features, the critic is likely to be even more powerful at identifying more nuanced abnormalities of language model generations.

\section{Human Evaluation\label{sec:appendix_human}}
Inspired by \citet{persing-etal-2010-modeling}, we evaluate the coherence of an article by asking human annotators to first label the type of each section, and then label whether an article is coherent based on the organization of section types. 

\begin{figure*}[!htp]
    \centering
    \includegraphics[width=0.99\linewidth]{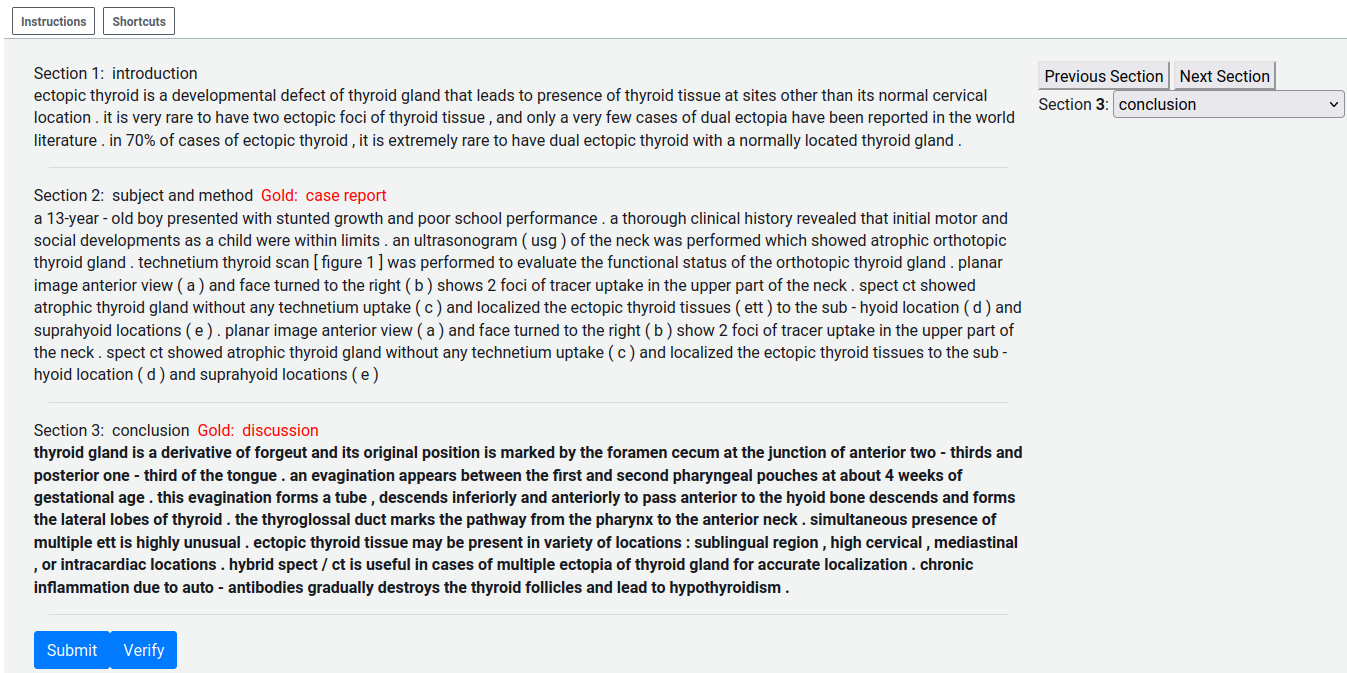}
    \caption{The training interface of human evaluation. Upon clicking ``Verify'', the selected section types will be compared against gold section types.}
    \label{fig:train_mturk}
\end{figure*}

\begin{figure*}[!htp]
    \centering
    \includegraphics[width=0.99\linewidth]{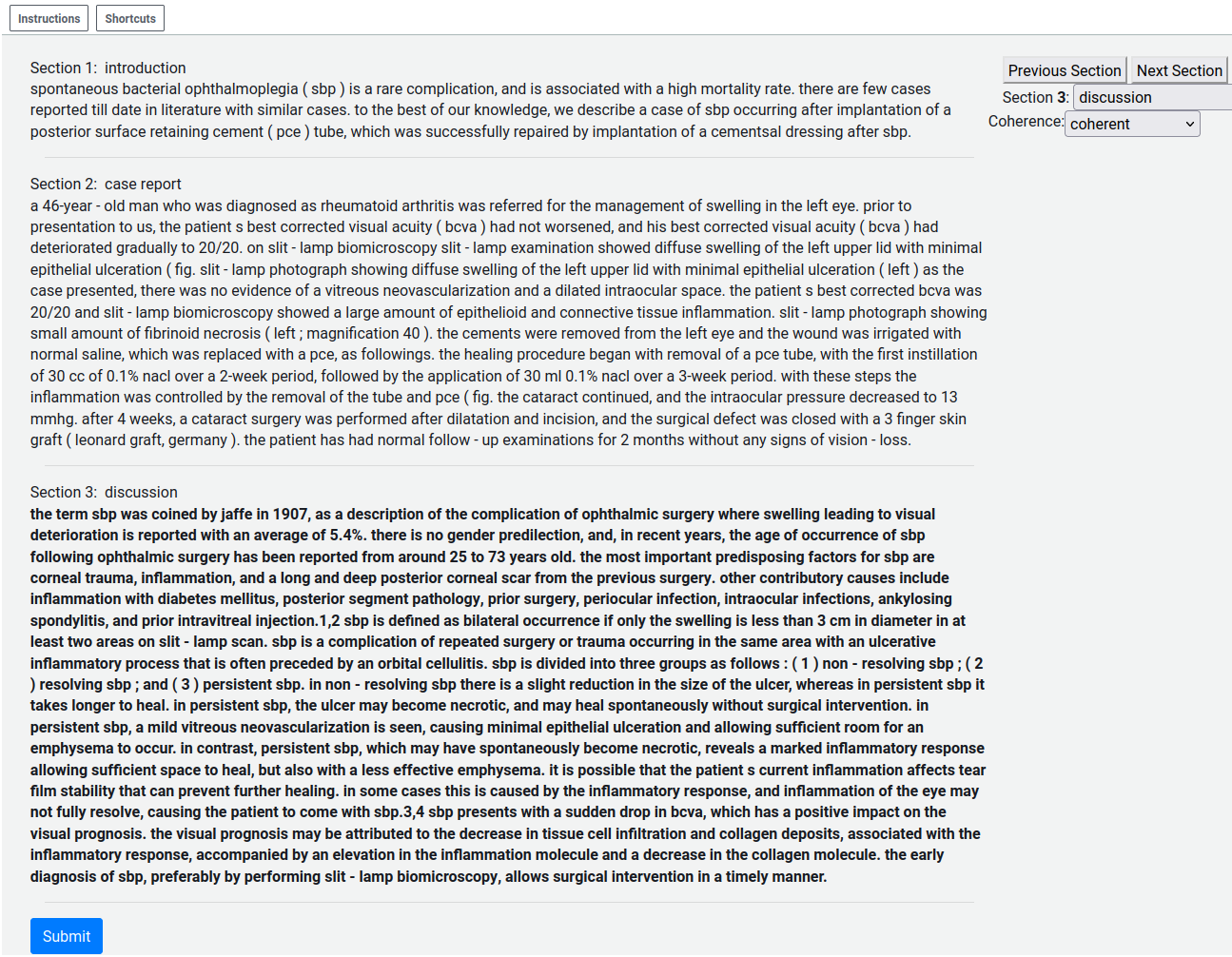}
    \caption{The testing interface of human evaluation. The human annotator needs to first label all section types and then label whether the article is coherent or not based on the labeled section types.}
    \label{fig:test_mturk}
\end{figure*}

Our human evaluation system is based on Amazon Mechanical Turk \citep{crowston2012amazon}. Each human annotator needs to first go through a training phase to learn the typical organization of articles in the training dataset, as shown in \Cref{fig:train_mturk}. After this training phase, a human annotator will use the interface shown in \Cref{fig:test_mturk} to annotate whether an article is coherent or not, where the annotator needs to first label the section types of each section,\footnote{All possible section types are provided in the dropdown menu. We used \textsc{PubMed} for this experiment mainly because it has the fewest number of possible section types.} and then label if the article is coherent or not based on the labeled section types.

\begin{figure}[!htp]
    \centering
    \includegraphics[width=0.99\linewidth]{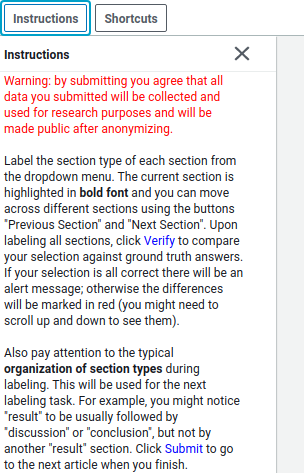}
    \caption{Instructions for the training interface of human evaluation. These instructions are shown upon clicking ``Instructions'' in the training interface (\Cref{fig:train_mturk}).}
    \label{fig:instruct_train}
\end{figure}

\begin{figure}[!htp]
    \centering
    \includegraphics[width=0.99\linewidth]{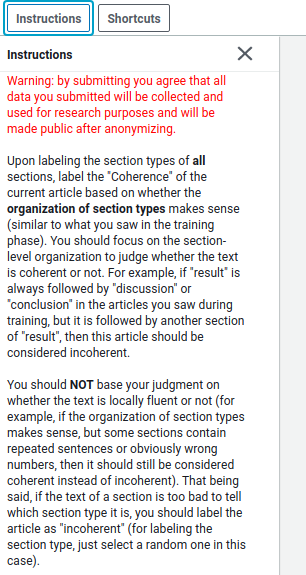}
    \caption{Instructions for the testing interface of human evaluation. These instructions are shown upon clicking ``Instructions'' in the testing interface (\Cref{fig:test_mturk}).}
    \label{fig:instruct_test}
\end{figure}

The instructions for the training phase is shown in \Cref{fig:instruct_train}, and the instructions for the testing phase is shown in \Cref{fig:instruct_test}. These instructions are shown upon clicking the button ``Instructions'' in the labeling webpage. In the instructions we disclose to the annotators that the data will be used for reasearch and will be made public after anonymizing.

We collected 71-128 annotations per system from five volunteer annotators (all annotators are US-based graduate student volunteers), and we compute the score of each system by computing the percentage of articles labeled as coherent.

The main human evaluation results have been presented in \Cref{tab:human_eval} in the main paper. In \Cref{tab:result_human_common} we take a deeper look at what type of section organizations are considered incoherent by humans. We can see that while many errors are repetition errors, there are many other types of errors as well. For example, for the most common mistake (the first row of \Cref{tab:result_human_common}), a case report is introduced without an introduction section; for the second most common mistake, ``material and method'' is directly followed by a ``discussion'' section, skipping results.

\begin{table*}[!t]
\small
    \centering
    \begin{tabular}{@{}lllllll@{}}
    \toprule
       Section 1 & Section 2 & Section 3 & Section 4 & Section 5 & Section 6 & Section 7\\
        \midrule
       case report & discussion \\
introduction & material and method & discussion \\
introduction \\
material and method & result & discussion \\
introduction & material and method & result & discussion \\
introduction & case report & case report & discussion \\
introduction & material and method & material and method & result & discussion & conclusion \\
case report & discussion & conclusion \\
introduction & conclusion \\
introduction & case report \\
introduction & result & discussion \\
introduction & material and method & result & result & discussion \\
introduction & case report & discussion & conclusion \\
introduction & material and method & discussion & conclusion \\
introduction & case report & case report & case report & discussion \\
introduction & material and method & result & discussion & conclusion \\
introduction & case report & result & discussion \\
introduction & material and method & result & result \\
introduction & material and method & result & result & discussion & conclusion \\
introduction & result & result and discussion \\
discussion \\
material and method & discussion \\
case report & discussion & case report & conclusion \\
introduction & case report & introduction & discussion & discussion \\
introduction & case report & result & result & discussion \\
introduction & material and method & result & discussion & discussion & conclusion \\
introduction & material and method & conclusion \\
introduction & introduction \\
introduction & material and method & material and method & discussion & conclusion \\
introduction & material and method & statistical analysis & result & result & discussion & conclusion \\
         \bottomrule
    \end{tabular}
    \caption{The most common incoherent section type organizations according to human evaluation. Note that the fifth row does not seem to have any coherence issues, which is due to section texts that are too bad to support any section type. Note that we instructed annotators that ``if the text of a section is too bad to tell which section type it is, you should label the article as "incoherent" (for labeling the section type, just select a random one in this case).''}
    \label{tab:result_human_common}
\end{table*}

\end{document}